%% file: main.tex
\documentclass[letterpaper, 10 pt, conference]{ieeeconf}
\IEEEoverridecommandlockouts 
\title{\LARGE \bf Neural Network Verification in Control}

\author{Michael Everett
\thanks{The author is with the Aerospace Controls Laboratory at the Massachusetts Institute of Technology, {\tt\small{mfe@mit.edu}}. This work was supported by Ford Motor Company.}%
}

\input{preamble}

\begin{document}

\maketitle

\begin{abstract}%
Learning-based methods could provide solutions to many of the long-standing challenges in control.
However, the neural networks (NNs) commonly used in modern learning approaches present substantial challenges for analyzing the resulting control systems' safety properties.
Fortunately, a new body of literature could provide tractable methods for analysis and verification of these high dimensional, highly nonlinear representations.
This tutorial first introduces and unifies recent techniques (many of which originated in the computer vision and machine learning communities) for verifying robustness properties of NNs.
The techniques are then extended to provide formal guarantees of \textit{neural feedback loops} (e.g., closed-loop system with NN control policy).
The provided tools are shown to enable closed-loop reachability analysis and robust deep reinforcement learning.
\end{abstract}
\noindent{\textit{\textbf{Software}}--\url{https://github.com/mit-acl/nn\_robustness\_analysis}}

\section{Introduction}

Data-driven (or learning-based) methods offer many compelling advantages for addressing long-standing issues in the field of controls.
For instance, real environments are often highly dynamic, nonlinear, uncertain, and high-dimensional, each of which present major challenges for the design, verification and deployment of control systems.
\cite{lamnabhi2017systems} identifies opportunities to reduce the costs of modeling complicated systems and improve the control of large-scale networked systems through learning: ``In order to maintain verifiable high performance, future engineering systems will need to be equipped with on-line capabilities for active model learning and adaptation, and for model accuracy assessment.''

Despite the potential advantages of learning-based control, this tutorial focuses on a fundamental open question: \textbf{how can we certify the safety, performance, and robustness properties of learning machines?}
Autonomous vehicles, passenger aircraft, and surgical robots are examples of control systems that need safety guarantees, while robots that explore oceans or other planets require assurances that an expensive, one-time experiment will succeed.
There are many challenges with solving these NN verification problems.

For instance, real-world uncertainties (e.g., from sensor noise, imperfect state estimates, unknown initial conditions, adversarial attacks~\cite{Szegedy_2014}) often appear at the NN input and must be mapped through the NN to make a formal statement about the set of possible NN outputs.
The high dimensionality and nonlinear nature of NNs introduces technical challenges for quickly and accurately propagating uncertainty sets through the NN.
Fortunately, there is a growing body of literature on this topic, much of which began in the computer vision and machine learning communities.
\cref{sec:nn_analysis} introduces and unifies the literature into a framework that can be tailored for specific applications.

\begin{figure}[t]
    \centering
    \includegraphics[width=\linewidth,trim=0 200 200 100,clip]{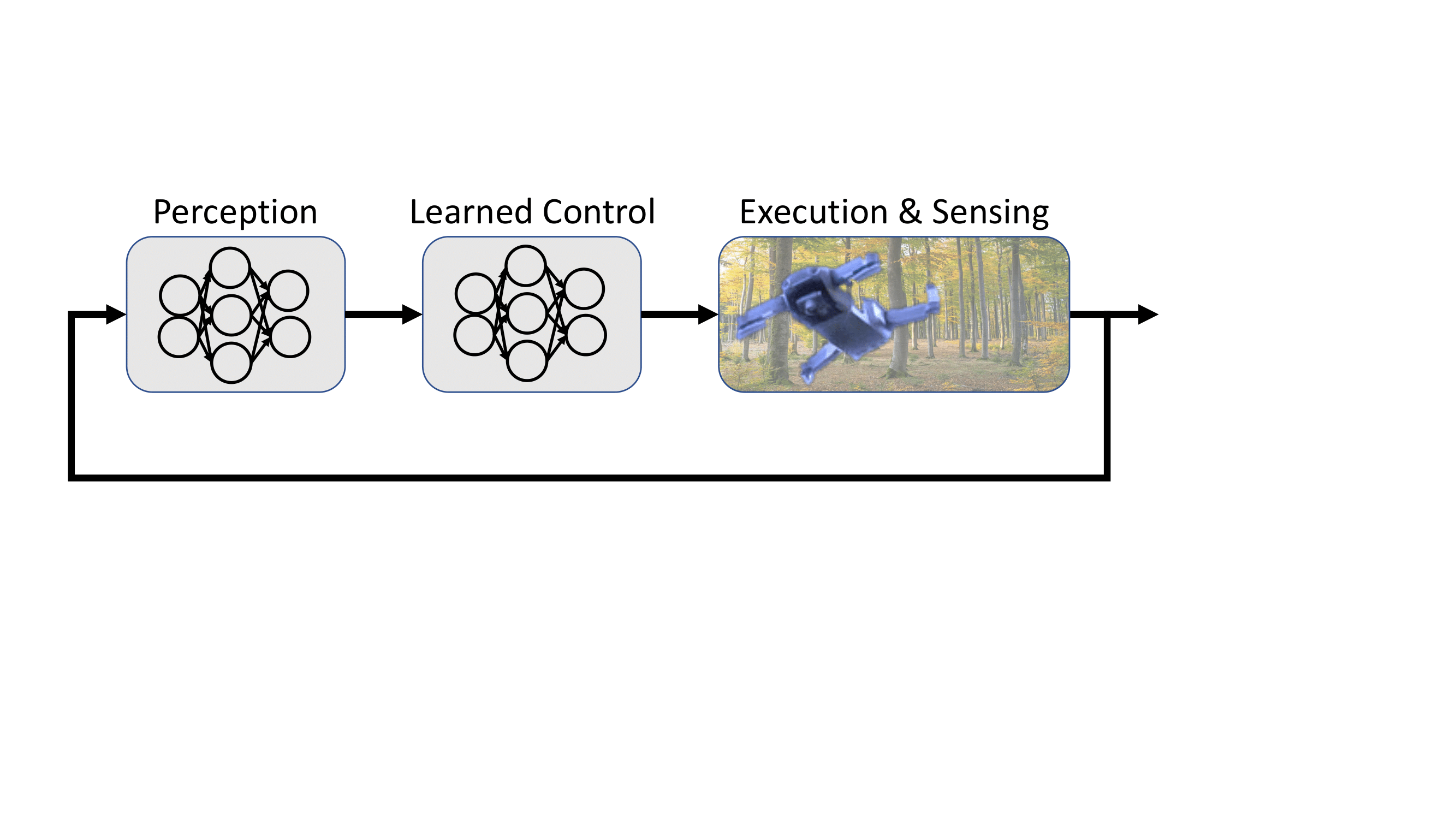}
    \caption{Neural Networks are a key component of many promising learning-based approaches for challenging controls problems. This tutorial introduces frameworks for the verification of robustness, safety, and performance properties, which is a critical step toward bringing learning-based methods to real-world control systems.}
    \label{fig:neural_feedback_loop_general}
\end{figure}

While NNs are difficult to analyze in isolation, the analysis is even more challenging for \textit{neural feedback loops} (NFLs) -- closed-loop systems with NN components, depicted in~\cref{fig:neural_feedback_loop_general}.
A fundamental challenge in NFLs is that NN outputs can influence the NN inputs at the next step.
Thus, conservatism about NN outputs compounds as time advances, which must be handled to verify meaningful properties.
\cref{sec:nfl_analysis} introduces the literature and describes a linear programming-based approach to estimate an NFL's \textit{forward reachable sets} -- the set of all states the closed-loop system could enter in some time interval.
The reachable sets could then be used to guarantee the control system will avoid undesirable states and terminate in a goal set.

In addition to verifying properties, NN analysis tools can be leveraged to address the so-called ``sim-to-real gap'' that plagues learning methods, and reinforcement learning in particular.
\cite{Huang_2017,behzadan2017vulnerability,yang2020enhanced,mandlekar2017adversarially,Rajeswaran_2017,Muratore_2018,Pinto_2017,morimoto2005robust,gleave2019adversarial,Uther_1997} provide examples where the performance of learned control policies can be degraded substantially by introducing a small uncertainties that were not present during training.
To address this issue with analysis tools, \cref{sec:robust_rl} introduces a method for robust reinforcement learning that considers the worst-case outcomes of its actions, with respect to known-to-be-imperfect state observations.

This tutorial is based on material in~\cite{everett2020robustness,everett2020certified,everett2021efficient,everett2021reachability} and provides further exposition and connections between the key problems.

\section{Literature Review}\label{sec:literature_review}

\subsection{Neural Networks in Control}

Early work on NNs in control systems dates back to the 1960s~\cite{widrow1964pattern,fu1970learning}, with substantial interest in the 1980s and 1990s, especially for the control of nonlinear systems~\cite{psaltis1988multilayered,li1989neural,narendra1991adaptive}.
Many of the NN architectures used today were developed in this era, including the convolutional neural network (CNN)~\cite{fukushima1982neocognitron} and long short-term memory (LSTM)~\cite{hochreiter1997long}.

So why are NNs a big deal again? What has changed?
One massive improvement is the ecosystem of tools: new computer hardware categories for faster training/inference (e.g., GPUs~\cite{sanders2010cuda}, TPUs~\cite{jouppi2017datacenter}), open-source software packages (e.g., Caffe~\cite{jia2014caffe}, TensorFlow~\cite{abadi2016tensorflow}, PyTorch~\cite{paszke2019pytorch}), massive curated datasets (e.g., ImageNet~\cite{deng2009imagenet}), and well-established benchmark simulation tasks (e.g., Atari~\cite{bellemare13arcade,mott1995stella}, Gym~\cite{Brockman_2016}).
In parallel, substantial algorithmic advances (e.g., dropout~\cite{srivastava2014dropout}), novel architectures (generative adversarial networks~\cite{goodfellow2014generative}, autoencoders~\cite{tschannen2018autoencoder}), and latency reductions (e.g., MobileNet~\cite{howard2017mobilenets}) expand what can be learned from data.

Another paradigm shift is the idea of \textit{deep} NNs (NNs with many hidden layers), as they unlock the ability to deal with high-dimensional inputs from modern perception systems (e.g., cameras).
Going far beyond the classical notions of state estimates from 1D sensors, DNNs provide a way to extract \textit{semantic knowledge} and \textit{context clues} in real-time from on-board sensing.
A body of work on \textit{end-to-end learning} takes this idea to an extreme by learning a mapping from raw sensor data to control.

Building on these advances, recent work demonstrates that NNs are a powerful representation for various control problems, including:
\begin{itemize}
	\item learning a control policy \cite{pomerleau1989alvinn,levine2014learning,tan2018sim}
	\item learning the system's model from data \cite{bansal2017hamilton}, possibly using physical constraints \cite{lutter2018deep,cranmer2020lagrangian,raissi2019physics,rackauckas2020universal}
	\item online estimation of states or parameters \cite{zhan2006neural,tagliabue2020touch}
	\item learning a representation for verification \cite{noroozi2008generation,yeung2019learning,bansal2020deepreach,chen2021learning}
\end{itemize}

Despite this progress, NNs still present many challenges for control systems.
For example, deep learning algorithms typically require massive amounts of training data, which can be expensive or dangerous to acquire and annotate.
NNs' lack of interpretability and possible over-confidence in unsupported predictions are common concerns, as well.
The combination of physics-based models with data is a promising approach, but there is not yet a consensus in the community on what should be modeled versus learned.
This tutorial focuses on the challenges of verifying systems that employ NNs.

\subsection{Verification in Controls}

Control design is primarily concerned with stability, performance, and robustness.
While there are many mathematical tools to aid the design of controllers and proof of these properties, there is usually still a gap between the models used for control design and the real, deployed systems.
Thus, \textit{formal verification} techniques provide assurances that the real system will meet specifications.
\cite{luckcuck2019formal} provides a recent surveys on the topic in the context of robotics.
\cite{garoche2019formal} and \cite{tabuada2009verification} provide further depth in verifying the software and hybrid nature of real control system implementations, respectively.
\cite{zheng2015perceptions} provides a survey on practitioners' perception of the state-of-the-art for these methods.

A main focus of this tutorial is \textit{reachability analysis}, which is a standard component of safety verification.
Informally, these methods compute the set of all states the system could ever reach, which can be used to provide guarantees that the system will, for example, reach the desired state and avoid dangerous regions of the state space.
Modern methods include Hamilton-Jacobi Reachability methods~\cite{tomlin2000game,bansal2017hamilton}, SpaceEx~\cite{frehse2011spaceex}, Flow*~\cite{chen2013flow}, CORA~\cite{althoff2015introduction}, and C2E2~\cite{duggirala2015c2e2,fan2016automatic}.
Orthogonal approaches that do not explicitly estimate the system's forward reachable set, but provide other notions of safety that assist in verification, include Lyapunov function search~\cite{papachristodoulou2002construction} and control barrier functions (CBFs)~\cite{ames2016control}.
Several recent methods learn a representation that aids in verification~\cite{noroozi2008generation,yeung2019learning,bansal2020deepreach,chen2021learning}.

Some of the key open issues include scalability to high-dimensional systems, computation time required for verification, and over-conservatism that can lead to arbitrarily loose performance bounds.
Moreover, a relatively small subset of the literature focuses on NN control policies, which present additional challenges as described next.

\subsection{Verification of Neural Feedback Loops}

Despite the importance of analyzing closed-loop behavior, much of the recent work on formal NN analysis has focused on NNs in isolation (e.g., for image classification)~\cite{Ehlers_2017, Katz_2017, Huang_2017b,Lomuscio_2017,Tjeng_2019,Gehr_2018}, with an emphasis on efficiently relaxing NN nonlinearities~\cite{gowal2018effectiveness,Weng_2018,singh2018fast,zhang2018efficient,Wong_2018,raghunathan2018certified,fazlyab2020safety}.
\cite{xu2020automatic} provides an excellent software repository for general computation graphs.
When NNs are embedded in feedback loops, a new set of challenges arise for verification, surveyed in~\cite{tran2020verification,xiang2018verification}.
To verify stability properties, \cite{tanaka1996approach} uses linear differential inclusions (LDI), and \cite{yin2021stability} uses interval constraint programming (ICP).

A handful of recent works~\cite{dutta2019reachability,huang2019reachnn,fan2020reachnn,ivanov2019verisig,xiang2020reachable,hu2020reach} propose methods that compute forward reachable sets of closed-loop systems with NN controllers.
A key challenge is in maintaining computational efficiency while still providing tight bounds on the reachable sets.
\cite{dutta2019reachability,huang2019reachnn,fan2020reachnn} use polynomial approximations of NNs to make the analysis tractable.
Most works consider NNs with ReLU approximations, whereas \cite{ivanov2019verisig} considers sigmoidal activations.
\cite{xiang2020reachable,yang2019efficient} introduce conservatism by assuming the NN controller could output its extreme values at every state.
\cite{vincent2021reachable} developed a method for ReLU NNs that also computes backward reachable sets, and \cite{hu2020reach} recently approached the problem with semidefinite programming (SDP).

Key open issues include the analysis of NFLs with high-dimensional, nonlinear, or stochastic plants and accounting for realistic perception systems (e.g., camera, lidar data).
Additionally, verification methods are rarely embedded in the learning process, which could be a path toward synthesizing known-to-be-safe NFLs, building on ideas from work on isolated NNs~\cite{gpu_implementation_crown}.

\input{nn_analysis}
\input{nfl_analysis}
\input{robust_rl}

\section{Discussion \& Open Challenges}

This tutorial provides an introduction to a collection of tools for analyzing and robustifying NNs for control systems.
After summarizing some of the relevant literature, the tutorial described a suite of methods for analyzing NNs in isolation.
Then, these methods were extended for closed-loop reachability analysis.
Finally, the analysis ideas were brought to deep RL to improve a learned policy's robustness to noise or adversarial attacks.

Each of the methods comes with its own limitations that could be the basis for future research.
For NN analysis, partitioning the input set provides a way of handling large uncertainties in a few dimensions, but does not scale particularly well to input sets with uncertainty in many dimensions.
It is likely that fundamentally different techniques are necessary in those cases.

For NFL analysis, reachable sets are one axis of safety verification, but other properties, such as stability and performance remain less studied.
The computational and conservatism challenges remain, and it is unclear how well existing methods extend to high-dimensional, nonlinear, and highly uncertain systems.
In addition, a perfect dynamics model may not be available, which presents another challenge for practical use of the method described in~\cref{sec:nfl_analysis}.

For robust RL, can the robustness analysis be extended to continuous action spaces?
When $\bm{\epsilon}_\text{adv}$ is unknown, as in our real systems, how could $\bm{\epsilon}_\text{rob}$ be tuned efficiently online while maintaining robustness guarantees?
How can the online robustness estimates account for other uncertainties (e.g., states that were not sufficiently explored in the training process) in a guaranteed manner?

More broadly, the topic of synthesizing known-to-be-safe and known-to-be-robust control policies using learning-based methods is a rich area for future study.
Toward that goal, this tutorial aims to provide an introduction to techniques for neural network verification in control.

\section*{Acknowledgment}
This tutorial is based on material in~\cite{everett2020robustness,everett2020certified,everett2021efficient} developed in collaboration with Golnaz Habibi, Bj\"{o}rn L\"{u}tjens, and Jonathan P. How.

\bibliographystyle{ieeetr} 
\bibliography{refs}

\end{document}

%% file: preamble.tex
\usepackage[utf8]{inputenc}
\usepackage{amsmath}
\usepackage{graphicx}
\usepackage{bm}
\usepackage{dsfont}
\usepackage{eufrak}
\usepackage{subcaption}
\usepackage{wrapfig}
\usepackage{epigraph}
\usepackage[dvipsnames]{xcolor}
\usepackage[sort,compress]{cite}
\usepackage{amsfonts}
\usepackage{url}
\usepackage{stmaryrd}

\usepackage[pdftex, pdfstartview={FitV}, pdfpagelayout={TwoColumnLeft},bookmarksopen=true,plainpages = false, colorlinks=true, linkcolor=black, citecolor = black, urlcolor = black,filecolor=black , pagebackref=false,hypertexnames=false, plainpages=false, pdfpagelabels ]{hyperref}
\usepackage{balance}
\usepackage[capitalize]{cleveref}

\newtheorem{theorem}{Theorem}[section]

\newtheorem{lemma}[theorem]{Lemma}

\newtheorem{definition}[theorem]{Definition}

\crefname{section}{Section}{Sections}
\crefname{theorem}{Theorem}{Theorems}
\crefname{lemma}{Lemma}{Lemmas}
\crefname{table}{Table}{Tables}
\crefformat{equation}{(#2#1#3)}

\newcommand{\Aout}{\mathbf{A}^\text{out}}
\newcommand{\bout}{\mathbf{b}^\text{out}}
\newcommand{\Ain}{\mathbf{A}^\text{in}}

\newcommand{\x}{\mathbf{x}}

\newcommand{\z}{\mathbf{z}}
\newcommand{\y}{\mathbf{y}}

\newcommand{\R}{\mathds{R}}

\newcommand{\istate}{k}
\newcommand{\ifacet}{i}
\newcommand{\icontrol}{j}

\newcommand{\piclbu}{\bm{\Gamma}}

\newcommand{\piclAu}{\bm{\Upsilon}}

\newcommand{\mout}{m_\text{out}}

\newcommand{\Xt}{\mathcal{X}_{t}}

\newcommand{\Xtt}{\mathcal{X}_{t+1}}

\newcommand{\xinX}{\mathbf{x}_t \in \Xt}

\newcommand{\Nuj}{\mathbf{n}^U_{\ifacet}}
\newcommand{\Muj}{\mathbf{M}^U_{\ifacet,:}}

\newcommand{\Ainbin}{\mathbf{A}^\text{in}\mathbf{x}_t\leq \mathbf{b}^\text{in}}

\newcommand{\Aoutj}{\mathbf{A}_{\ifacet,:}^{\textrm{out}}}
\newcommand{\AoutjBt}{\Aoutj \mathbf{B}_{t}}

\newcommand{\AoutjBtPiclAuCt}{\Aoutj \mathbf{B}_{t} \piclAu_{\ifacet,:,:} \mathbf{C}_t^T}

\newcommand{\boutj}{\bout_{\ifacet}}

\newcommand{\xnom}{\mathring{\mathbf{x}}}

\newcommand{\xtnom}{\xnom_{t}}

\newcommand{\bpxnom}{\mathcal{B}_p(\xnom, \bm{\epsilon)}}

\newcommand{\binfxnom}{\mathcal{B}_{\infty}(\xnom, \epsilon)}
\newcommand{\bpxtnom}{\mathcal{B}_p(\xtnom, \epsilon)}

\newcommand{\xttbnd}{\bm{\gamma}}

\newcommand{\xttubistate}{\xttbnd^U_{t+1,\istate}}

\newcommand{\xttubifacet}{\xttbnd^U_{t+1,\ifacet}}

\newcommand{\CROWNAu}{\mathbf{\Psi}}
\newcommand{\CROWNAl}{\mathbf{\Phi}}
\newcommand{\CROWNbu}{\bm{\alpha}}
\newcommand{\CROWNbl}{\bm{\beta}}

\newcommand{\sinxnom}{S_\text{in}(\x^\text{nom})}

\newcommand{\Jbar}[3]{s(#1, #2, #3)}

\usepackage{accents}
\newcommand{\ubar}[1]{\underaccent{\bar}{#1}}

\DeclareMathOperator*{\argmax}{\arg\!\max}
\DeclareMathOperator*{\argmin}{\arg\!\min}

\usepackage{bibentry}
\nobibliography*

\usepackage[addedmarkup=bf]{changes}
\definechangesauthor[name={MFE}, color={blue}]{MFE}
\definechangesauthor[name={GH}, color={ForestGreen}]{GH}
\definechangesauthor[name={JH}, color={red}]{JH}

\usepackage[acronym]{glossaries}
\makeglossaries
\newacronym{rl}{RL}{Reinforcement Learning}
\newacronym{sl}{SL}{Supervised Learning}
\newacronym{dnn}{DNN}{Deep Neural Network}
\newacronym{dnns}{DNN}{Deep Neural Networks}
\newacronym{lstm}{LSTM}{Long Short-Term Memory}
\newacronym{carrl}{CARRL}{Certified Adversarial Robustness for Deep Reinforcement Learning}
\newacronym{orca}{ORCA}{Optimal Reciprocal Collision Avoidance}
\newacronym{relu}{ReLU}{Rectified Linear Unit}
\newacronym{dqn}{DQN}{Deep Q-Network}
\newacronym{lp}{LP}{Linear Programming}
\newacronym{smt}{SMT}{Satisfiability Modulo Theory}
\newacronym{fgst}{FGST}{Fast Gradient Sign method with Targeting}
\newacronym{crown}{CROWN}{Algorithm from~\cite{Weng_2018b}}
\newacronym{ga3c}{GA3C}{Hybrid GPU/CPU Asynchronous Advantage Actor-Critic}
\newacronym{cadrl}{CADRL}{Collision Avoidance with Deep Reinforcement Learning}
\newacronym{ga3ccadrl}{GA3C-CADRL}{\acrfull{ga3c} \acrshort{cadrl}}

\usepackage{multirow}

\newcommand{\citep}{\cite}
\usepackage{hhline}
\usepackage{empheq}
\usepackage[most]{tcolorbox}

\newenvironment{alignbox}{%
  \empheq{align}
}{\endempheq}

\tcolorboxenvironment{alignbox}{colback=yellow!20!white}

\usepackage[T1]{fontenc}

%% file: nn_analysis.tex

\newcommand{\robustnessproblem}{\cref{eqn:rob_analysis}}

\section{Analyzing NNs in Isolation}\label{sec:nn_analysis}

\begin{table*}
\begin{tabular}{|p{1.2cm}||p{4.9cm}|p{4.9cm}|p{4.9cm}|}
	\hline
	Problem & \textbf{Reachability} & \textbf{Verification} & \textbf{Minimal Adversarial Example} \\
	\hhline{|=#=|=|=|}
	Graphic &
	\begin{subfigure}{\linewidth}
		\vspace{0.1in}
		\centering
		\includegraphics[width=\linewidth, trim =0 830 1100 0, clip]{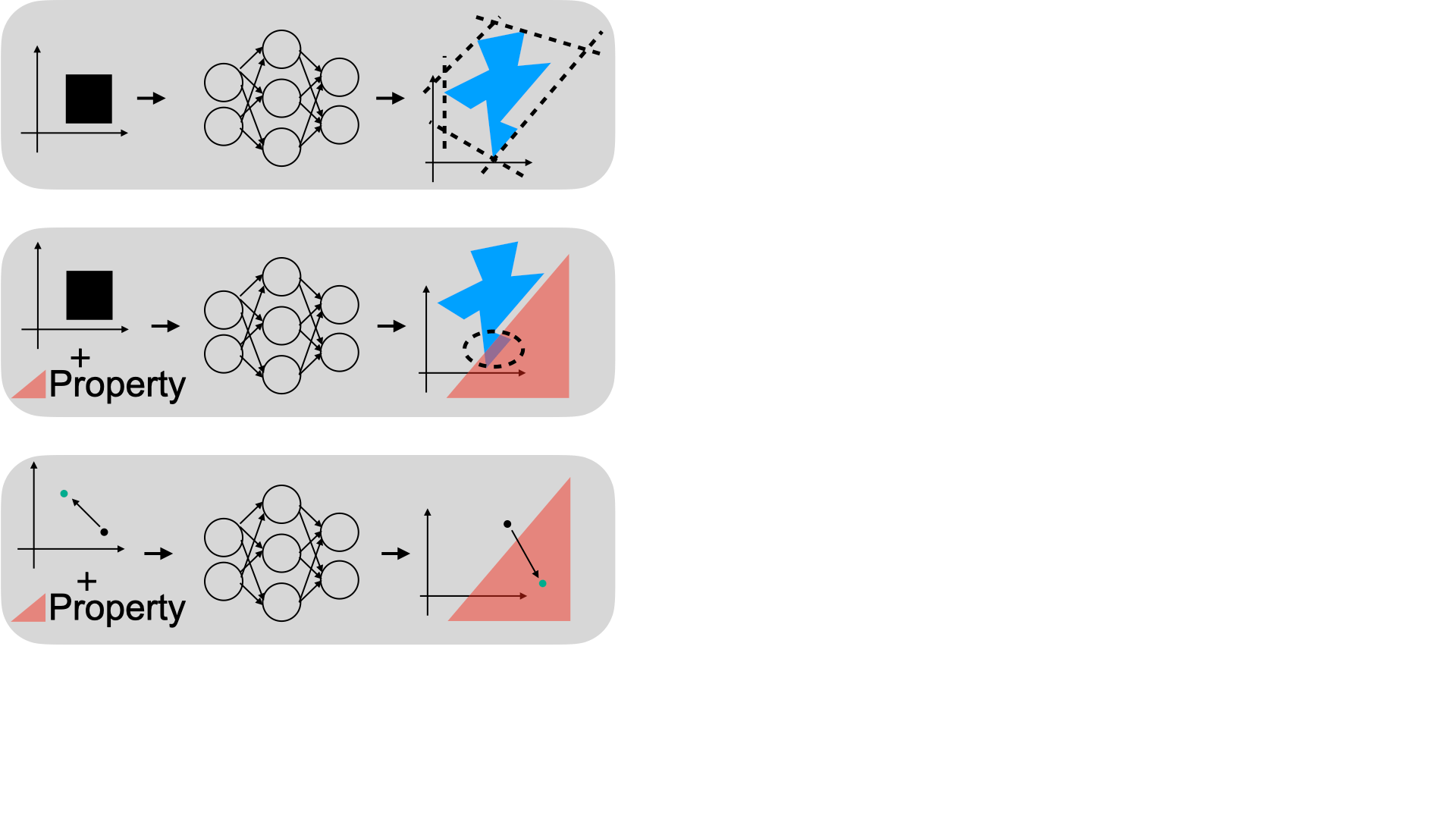}
		\label{fig:nn_analysis_reachability}
	\end{subfigure}
	&
	\begin{subfigure}{\linewidth}
		\vspace{0.1in}
		\centering
		\includegraphics[width=\linewidth, trim =0 530 1100 300, clip]{figures/robustness_analysis_problems.png}
		\label{fig:nn_analysis_verification}
	\end{subfigure}
	&
	\begin{subfigure}{\linewidth}
		\vspace{0.1in}
		\centering
		\includegraphics[width=\linewidth, trim =0 230 1100 600, clip]{figures/robustness_analysis_problems.png}
		\label{fig:nn_analysis_adv_example}
	\end{subfigure} 
	\\ 
	\hline
Problem & Compute NN's reachable set from $\sinxnom$, known formally as ``the image of $\sinxnom$ under the DNN $f$'', written as $f[\sinxnom]=\{ f(\x)\ \lvert\ \x\in \sinxnom \}$ 
&
Will NN classifier output same label for every $\x^{(0)}\in\sinxnom$?
Recall that in classification, NN predicts an input's label, $i^*=\argmax_{j} f_j(\x^\text{nom})$.
&
Unknown $\sinxnom$.
For classification, \textit{minimal adv. example} is result of applying smallest perturbation to nominal input $\x^\text{nom}$ to cause non-nominal predicted label.
\\
\hline
 Answer & Output Set & Yes/No/Unsure & Perturbed Input \\ 
 \hline 
Algorithm Sketch &
For $\ell_\infty$-ball output set, solve \robustnessproblem\ for $2n_{L+1}$ objectives,
{\setlength{\belowdisplayskip}{3pt} \setlength{\abovedisplayskip}{3pt}
\begin{align}
c_i(\z^{(L)})&=\mathrm{sign}(i)\cdot \mathbf{e}_i^T \left(\z^{(L)}\right) \nonumber \\
&\forall i\in \{-n_{L+1}, \ldots, n_{L+1}\}\setminus 0, \nonumber
\end{align}where ${\mathbf{e}_i\in\mathbb{R}^{n^{(L)}}}$ is standard basis vector.} 
  &
Solve \robustnessproblem\ for $n_{L+1}-1$ objectives (1 per non-nominal label), $c_i(\z^{(L)})=(\mathbf{e}_{i^*}-\mathbf{e}_i)^T \z^{(L)} \quad \forall i\in \mathcal{I}=[n_{L+1}-1]\setminus i^*$, and let $p_i$ be the cost of the solution $\forall i\in \mathcal{I}$.
If $p_i>0 \ \forall i\in \mathcal{I}$, then the network's robustness is \textit{verified} for the input $\x^\text{nom}$ and its neighborhood $\sinxnom$.
   &
Assume something about input set's structure, (e.g., $\sinxnom\equiv\ell_p$-ball with radius ${\bm{\delta}\in\mathbb{R}}$).
Change \cref{eqn:rob_analysis}'s objective to $\bm{\delta}$ and add constraint $f_{i^*}(\x^{(0)})-f_i(\x^{(0)})>0$.
Alternatively, use branch-and-bound~\cite{Weng_2018} (i.e., iteratively solve \textit{verification} (middle column) for various sizes of $S_\text{in}(\x^\text{nom})$). \\
\hline
\end{tabular}
\caption{Common Problems in NN Verification. The formulation in \robustnessproblem\ provides a framework for thinking about each one.}
\label{table:nn_analysis_problems}
\end{table*}

This tutorial begins by introducing tools for formally analyzing NNs (in isolation).
While empirical performance statistics can indicate that a NN has learned a useful input-output mapping, there are still concerns about how much confidence to associate with decisions resulting from a learned system.
One direction toward providing a confidence measure is to consider how the various sources of uncertainty in training/execution processes map to uncertainty in outputs of trained NNs.
Many of these uncertainties appear at the NN input (e.g., from noisy/adversarially attacked sensing, unknown initial conditions), thus this work focuses on the problem of propagating input uncertainties through NNs to bound the set of possible NN outputs online.


\subsection{NN Notation \& Analysis Problems}\label{sec:nn_analysis:nn_notation}

For a feedforward neural network with $L$ hidden layers (plus one input and one output layer), the number of neurons in each layer is $n_l\in[L+1]$, where $[i]$ denotes the set $\{0,1,\ldots,i\}$.
We denote the $l$-th layer weight matrix $\mathbf{W}^{(l)}\in\R^{n_{l+1}\times n_{l}}$, bias vector $\mathbf{b}^{(l)}\in\R^{n_{l+1}}$, and coordinate-wise activation function $\sigma^{(l)}: \R^{n_{l+1}} \to \R^{n_{l+1}}$, including ReLU (i.e., $\sigma(\mathbf{z})=\mathrm{max}(0,\mathbf{z})$), tanh, sigmoid, and many others.
For input $\x\in\R^{n_0}$, the NN output $f(\x)$ is,
\begin{align}
\begin{split}
	\x^{(0)} &= \x \\
	\z^{(l)} &= \mathbf{W}^{(l)} \x^{(l)}+\mathbf{b}^{(l)}, \forall l\in[L] \\
	\x^{(l+1)} &= \sigma^{(l)}(\z^{(l)}), \forall l\in[L-1] \\
    f(\x) &= \z^{(L)}.
\end{split}
\end{align}

Many NN analysis problems aim to say something about the network output given a set of possible network inputs.
These problems can be generalized with the following optimization problem~\cite{salman2019convex}:
\begin{alignbox}
\begin{split}
    \min_{\x^{(0)}\in \sinxnom} \quad & c(\z^{(L)}) \\
    \mathrm{s.t.} \quad & \z^{(l)}=\mathbf{W}^{(l)}\x^{(l)}+\mathbf{b}^{(l)}, \forall l\in[L] \\ 
    & \x^{(l+1)} = \sigma^{(l)}(\z^{(l)}), \forall l\in[L-1] \label{eqn:rob_analysis}
\end{split}
\end{alignbox}
\noindent which minimizes a property of the NN output, $c(\z^{(L)})$, subject to the constraints that the input must lie within some set $S_\text{in}(\x^\text{nom})$ around a nominal input $\x^\text{nom}$.
The \textit{specification} function: $c: \mathbb{R}^{n_L} \to \mathbb{R}$ defines which property of the NN output is  analyzed.

Three examples, reachability, verification, and minimal adversarial example problems, are visualized and briefly summarized in~\cref{table:nn_analysis_problems}.
Note that each of these problems solves \robustnessproblem, or a slightly modified version.

\subsection{NN Analysis Methods}

Analysis of how a set of possible inputs propagates through a NN has an inherent tradeoff between computation time and conservatism.
This tradeoff is a result of two main issues in the optimization problem \robustnessproblem: the nonlinear activations $\sigma$ and the sets $\sinxnom$.
First, the nonlinear activations $\sigma$ make the problem non-convex, which greatly reduces the set of problems that can be solved exactly.
In the special case of ReLU activations (which are piecewise linear), the \robustnessproblem \textit{can} be solved exactly with mixed integer programming, but the problem is still NP-Complete~\cite{katz2017reluplex,weng2018towards} and the solve times are often prohibitively slow for reasonably sized NNs.
Thus, most of the literature focuses on how to \textit{relax} the nonlinearities in the last constraint of \cref{eqn:rob_analysis}, such that bounds on \robustnessproblem\ can be computed with efficient convex optimization techniques (e.g., linear, quadratic, semidefinite programming).
These relaxations lead to the second issue, namely that convex relaxations of a nonlinearity become less and less tight as the size of $\sinxnom$ increases.



\begin{figure*}[t]
	\centering
	\includegraphics[width=\linewidth, trim=0 850 0 0, clip]{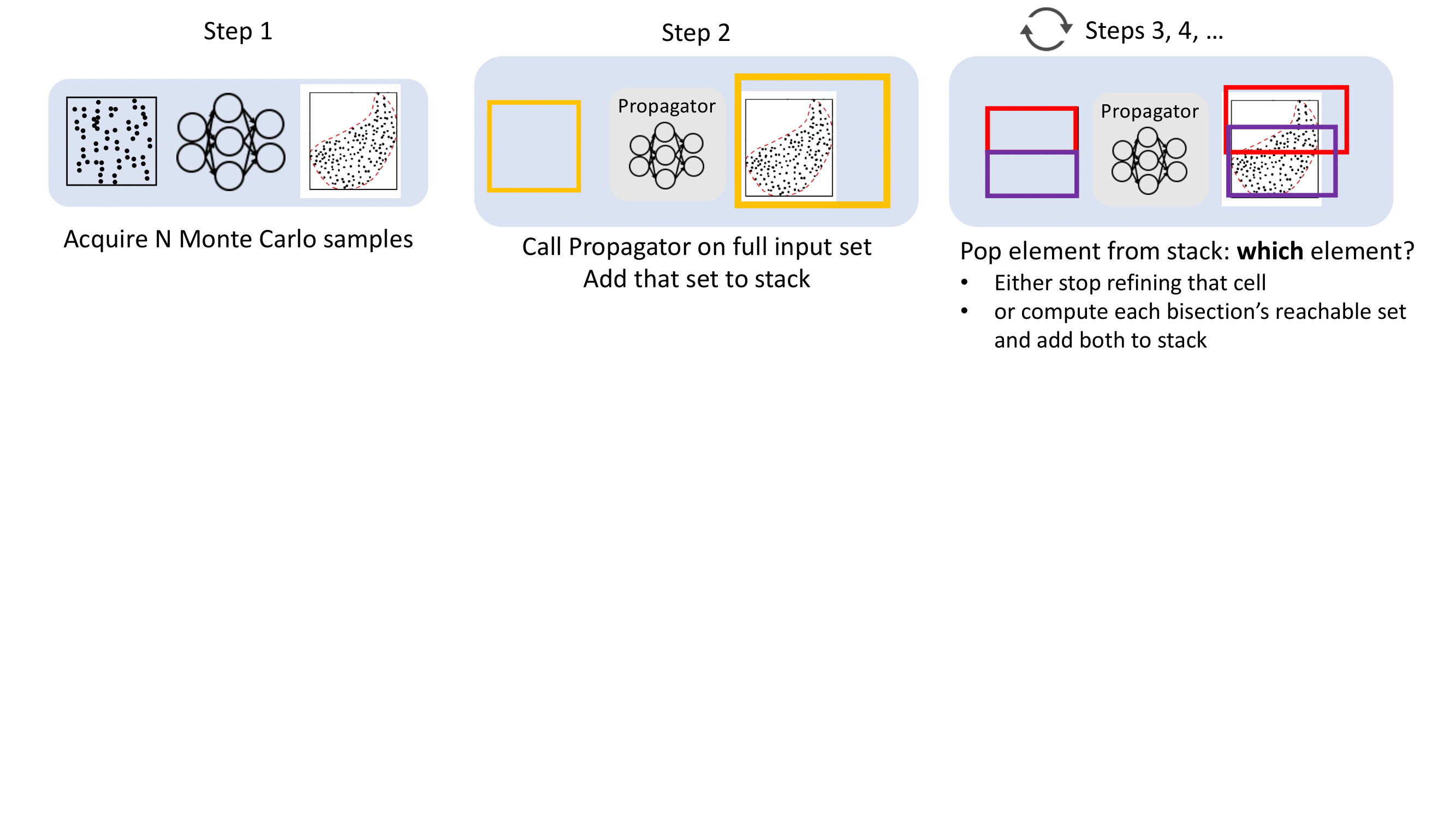}
	\caption{
		Partitioner \& Propagator Framework. The \textit{Partitioner} determines how to split the input set (here, a Sim-Guided approach based on Monte Carlo sampling is illustrated~\cite{xiang2020reachable,everett2020robustness}), while the \textit{Propagator} maps sets through the trained NN.
	}
	\label{fig:nn_analysis_algorithm}
	\vspace*{-.1in}
\end{figure*}

To address these two challenges, there are two separate paradigms in the literature.
\textit{Propagators} estimate how the full input set moves through the network, and they primarily differ in approximation strategies of the nonlinear activation functions.
At one extreme, Interval Bound Propagation (IBP)~\cite{gowal2018effectiveness} approximates the output of each layer with an $\ell_\infty$ ball, leading to conservative but fast-to-compute bounds of the final layer.
Linear relaxation-based techniques~\cite{salman2019convex,xu2020automatic} often achieve tighter bounds with more computation by approximating nonlinear activations with linear bounds -- some of these can be solved in closed form~\cite{Weng_2018,zhang2018efficient}.
Other propagators provide tighter analysis at the cost of higher computation time, including approaches based on QP/SDP~\cite{raghunathan2018certified,fazlyab2020safety,dathathri2020enabling}, and convex relaxation refinements (e.g., jointly relaxing multiple neurons~\cite{singh2019beyond,tjandraatmadja2020convex,muller2021prima}).

For example, CROWN~\cite{zhang2018efficient} computes affine bounds on the NN output using the following result.
\begin{theorem}[From~\cite{zhang2018efficient}, Convex Relaxation of NN]\label{thm:crown_particular_x}
Given a NN $f:\R^{n_0}\to\R^{n_m}$ with $m-1$ hidden layers, there exist two explicit functions $f_j^L: \R^{n_0}\to\R^{n_m}$ and $f_j^U: \R^{n_0}\to\R^{n_m}$ such that $\forall j\in [n_m-1], \forall \mathbf{x}\in\mathcal{B}_p(\mathbf{x}_0, \epsilon)$, the inequality $f_j^L(\mathbf{y})\leq f_j(\mathbf{x})\leq f_j^U(\mathbf{x})$ holds true, where
\begin{align}
\label{eq:f_j_UL}
    f_{j}^{U}(\x) &= \CROWNAu_{j,:} \x + \CROWNbu_j \\
    f_{j}^{L}(\x) &= \CROWNAl_{j,:} \x + \CROWNbl_j,
\end{align}
where $\CROWNAu, \CROWNAl \in \R^{n_m \times n_0}$ and $\CROWNbu, \CROWNbl \in \R^{n_m}$ are defined recursively using NN weights, biases, and activations (e.g., ReLU, sigmoid, tanh), as detailed in~\cite{zhang2018efficient}.
After relaxing the nonlinear NN to affine bounds within some input region, it is possible to maximize/minimize the relaxed NN's output in closed form.
\end{theorem}

\textit{Partitioners} break the input set into smaller regions, compute the reachable set of each small region, and return the total reachable set as the union of each smaller region's reachable set.
The key difference between partitioning approaches is the strategy for how to split the input set.
Some works make one bisection of the input set~\cite{anderson2020tightened}; \cite{xiang2018output} splits the input set into a uniform grid;
\cite{wang2018formal} uses gradients to decide which cells to split for ReLU NNs.
\cite{rubies2019fast} improves on~\cite{wang2018formal} using ``shadow prices'' to optimize \textit{how} to split a particular cell (i.e., along which dimension), but does not provide a way of choosing \textit{which} cells to split when computing tight reachable sets.
As illustrated in~\cite{xiang2020reachable}, substantial performance improvements can be achieved by stopping the refinement of cells that are already sufficiently refined.
Thus, the Simulation-Guided approach (SG)~\cite{xiang2020reachable}, uses a partitioning strategy where Monte Carlo samples of the exact NN output are used as guidance for efficient partitioning of the input set, reducing the amount of computation required for the same level of bound tightness.

The two paradigms of Propagators and Partitioners have been developed separately toward a similar objective (tightly analyzing NNs).
For instance, SG used IBP to compute output sets.
In this tutorial, we describe how the two paradigms can be unified.
In~\cite{everett2020robustness}, we show that improvements in robustness analysis lead to reduced conservatism in control tasks.

\subsection{Approach}

\begin{figure}[t]
	\centering
	\includegraphics[width=\linewidth, trim=0 70 0 40, clip]{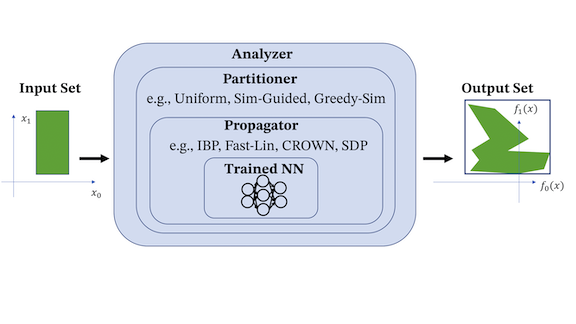}
	\caption{
		Robustness Analysis Architecture. This tutorial shows how Partitioners and Propagators can be unified to efficiently bound the set of NN outputs for a given input set.
	}
	\label{fig:architecture}
	\vspace*{-.1in}
\end{figure}

\cref{fig:architecture} shows a schematic of the proposed framework with its three nested modules: Analyzer, Partitioner and Propagator.
The Analyzer is aware of the desired output shape (e.g., lower bounds, $\ell_\infty$-ball, convex hull) and termination condition (e.g., computation time, number of Propagator calls, improvement per step).
The Analyzer specifies a Propagator (e.g., CROWN~\cite{zhang2018efficient}, IBP~\cite{gowal2018effectiveness}, SDP~\cite{fazlyab2020safety}, Fast-Lin~\cite{Weng_2018}) and a Partitioner (e.g., Uniform~\cite{xiang2018output}, Sim-Guided~\cite{xiang2020reachable} or the algorithms proposed in this section).
The Partitioner decides how to split the input set into cells, and the Propagator is used by the Partitioner to estimate the output set corresponding to an input set cell.

As illustrated in~\cref{fig:nn_analysis_algorithm}, the SG~\cite{xiang2020reachable} partitioning algorithm tightens IBP's approximated boundary with the following key steps:
(1) acquire $N$ Monte Carlo samples of the NN outputs to under-approximate the reachable set as the interval $[u_{\text{sim}}]$ (where $\llbracket \cdot \rrbracket$ denotes a closed $n$-dimensional interval), (2) using IBP, compute the reachable set of the full input set and add this set to a stack $M$, and (3) (iteratively) pop an element from $M$, and either stop refining that cell if its computed reachable set is within $\llbracket u_\text{sim} \rrbracket$, or bisect the cell, compute each bisection's reachable set, and add both to the stack.
The SG algorithm terminates when one of the cell's dimensions reaches some threshold, and the returned reachable set estimate is the weighted $\ell_\infty$-ball that surrounds the union of all of the cells remaining on the stack and $\llbracket u_{\text{sim}} \rrbracket$.

\cite{everett2020robustness} proposes a partitioning algorithm with better bound tightness for the same amount of computation, called Greedy-Sim-Guided (GSG), by modifying the choice of which cell in $M$ to refine at each step.
Rather than popping the first element from the stack (LIFO) as in SG, GSG refines the input cell with corresponding output range that is furthest outside the output boundary of the $N$ samples.
Whereas SG might choose a cell that is not pushing the overall boundary outward at a given iteration, GSG will always choose to refine an input cell that is pushing the boundary.
This heuristic gives the opportunity to reduce the boundary estimate at each iteration.
While the core SG algorithm remains the same, the greedy strategy can greatly improve the algorithm's performance.

GSG also optimizes for the desired output shape.
For example, if the goal is to find a convex hull over-approximation, GSG pops the input set from the stack with the output set that is furthest from the convex hull boundary (instead of the $\ell_\infty$-ball, as in SG~\cite{xiang2020reachable}).

Because the number of partitions can scale exponentially with the input dimension in the worst case (e.g., to split each input dimension in half, $n_\text{partitions}=2^{n_0}$), partition-based refinement is most effective for problems where either the NN's input vector is small or the input uncertainty is only non-zero on a handful of the input dimensions.
In the context of control systems, this could correspond to a NN controller for a system with either a small state/observation space, or a large state space with uncertainty only on some states.
This open challenge motivates future work to project high-dimensional NN inputs onto lower-dimensional spaces, or on new types of refinement methods designed for high-dimensional inputs.
Note that propagators generally scale well with input dimension, as many approaches have been demonstrated on image classification NNs.

\begin{figure*}[t]
\centering
\begin{subfigure}{0.33\linewidth}
	\centering
	\includegraphics[width=\textwidth, trim=0 0 0 80, clip]{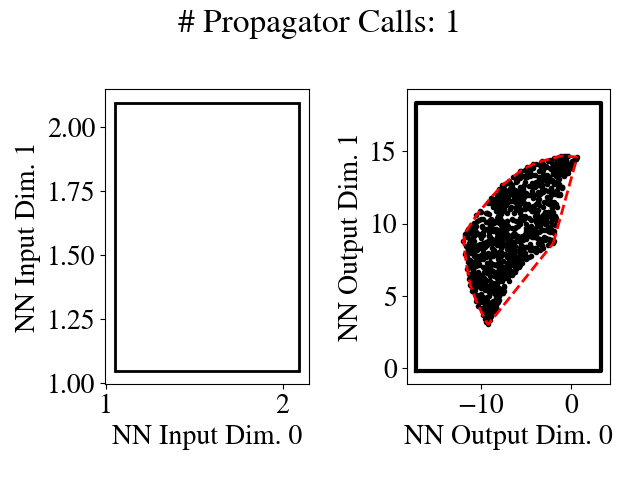}
	\caption{Iteration 0~\cite{zhang2018efficient} (1 Propagator Call)}
	\label{fig:nn_analysis_animation:0}
\end{subfigure}%
\begin{subfigure}{0.33\linewidth}
	\centering
	\includegraphics[width=\textwidth, trim=0 0 0 80, clip]{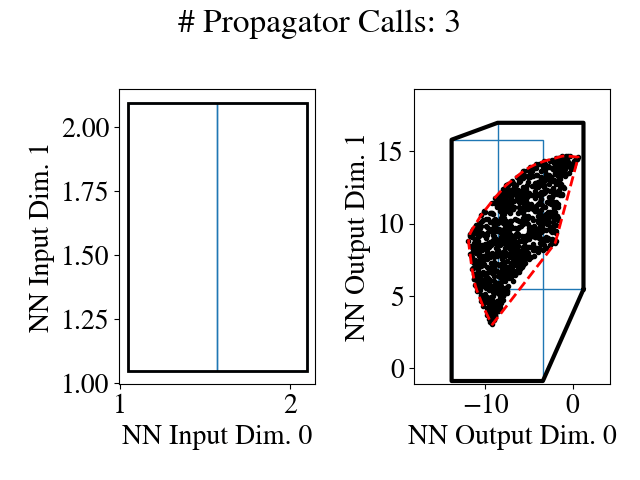}
	\caption{Iteration 1 (3 Propagator Calls)}
	\label{fig:nn_analysis_animation:1}
\end{subfigure}%
\begin{subfigure}{0.33\linewidth}
	\centering
	\includegraphics[width=\textwidth, trim=0 0 0 80, clip]{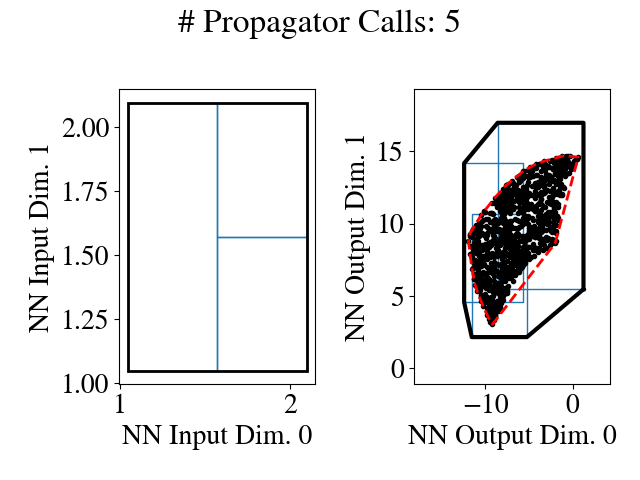}
	\caption{Iteration 2 (5 Propagator Calls)}
	\label{fig:nn_analysis_animation:2}
\end{subfigure}
\\ \vspace{0.1in}
\begin{subfigure}{0.33\linewidth}
	\centering
	\includegraphics[width=\textwidth, trim=0 0 0 80, clip]{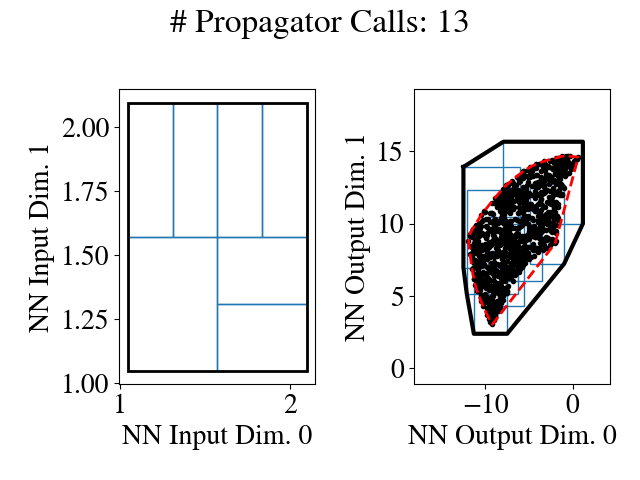}
	\caption{Iteration 5 (13 Propagator Calls)}
	\label{fig:nn_analysis_animation:5}
\end{subfigure}%
\begin{subfigure}{0.33\linewidth}
	\centering
	\includegraphics[width=\textwidth, trim=0 0 0 80, clip]{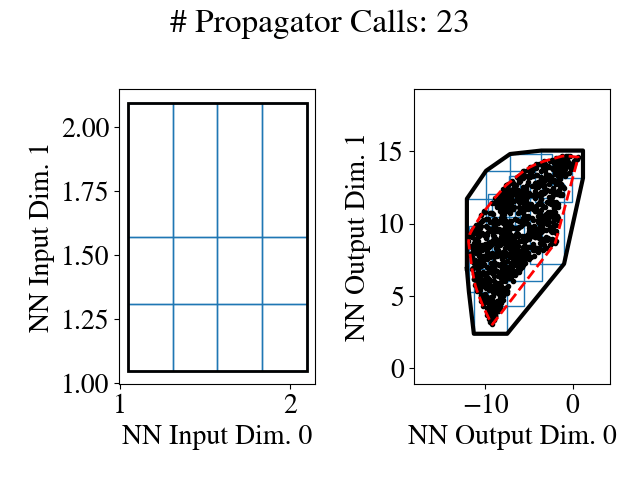}
	\caption{Iteration 10 (23 Propagator Calls)}
	\label{fig:nn_analysis_animation:10}
\end{subfigure}%
\begin{subfigure}{0.33\linewidth}
	\centering
	\includegraphics[width=\textwidth, trim=0 0 0 80, clip]{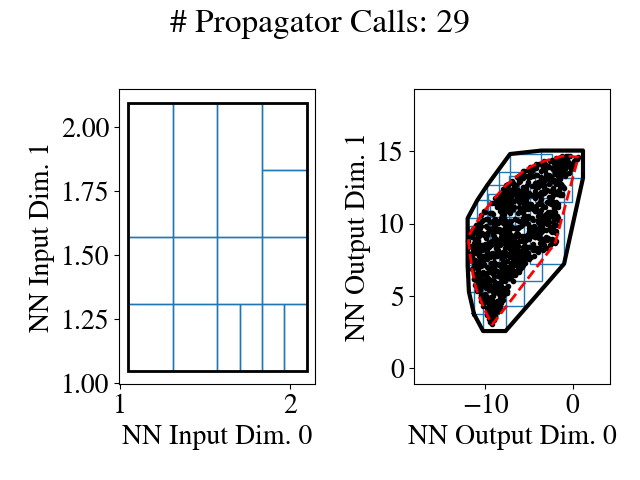}
	\caption{Iteration 14 (29 Propagator Calls)}
	\label{fig:nn_analysis_animation:14}
\end{subfigure}%
\caption{
GSG Partitioner~\cite{everett2020robustness} with CROWN Propagator~\cite{zhang2018efficient}. Note that (a) is equivalent to CROWN. As the input set is refined (b)-(f), the estimated bound (black, over-approximation) gets closer to the convex hull around the MC samples (dashed red, under-approximation).
}
\label{fig:nn_analysis_animation}
\end{figure*}

\subsection{Results}

\cref{fig:nn_analysis_animation} shows how a Partitioner can tighten the bounds provided by a Propagator.
Using the CROWN Propagator~\cite{zhang2018efficient}, (a) shows the relatively conservative bounds for a small NN from~\cite{xiang2020reachable}.
Moving to the right (b)-(f), the input set is refined according to the GSG Partitioner~\cite{everett2020robustness}, and the output set bounds become tighter.

\begin{figure}[t]
\centering
\begin{subfigure}{0.33\linewidth}
	\centering
	\includegraphics[width=\textwidth, trim=0 0 0 0, clip]{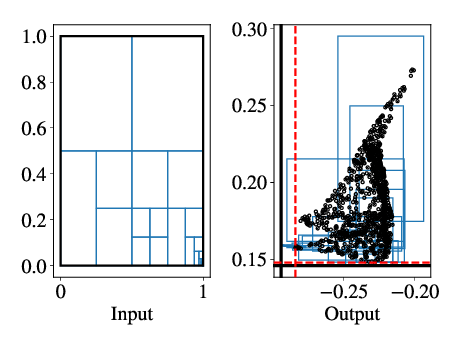}
	\caption{ Lower Bounds}
	\label{fig:tightest_lower_bounds}
\end{subfigure}%
\begin{subfigure}{0.33\linewidth}
	\centering
	\includegraphics[width=\textwidth, trim=0 0 0 0, clip]{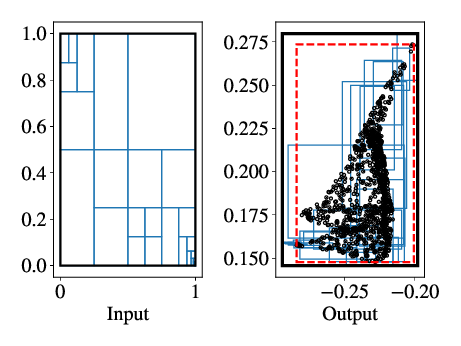}
	\caption{ $\ell_{\infty}$-ball}
	\label{fig:tightest_linf_ball}
\end{subfigure}%
\begin{subfigure}{0.33\linewidth}
	\centering
		\includegraphics[width=\textwidth, trim=0 0 0 0, clip]{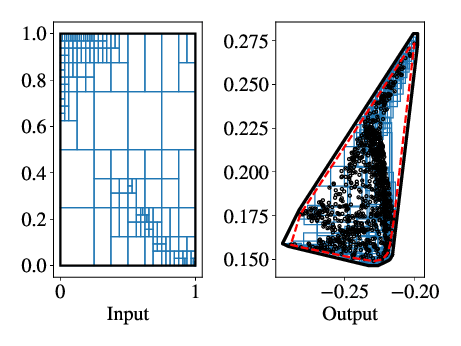}
	\caption{ Convex Hull}
	\label{fig:tightest_convex_hull}
\end{subfigure}%
\caption{\label{fig:boundary}
Input \& output sets for different output set shapes.
The estimated bounds (black) are ``tight'' when they are close to the bounds from exhaustive sampling (dashed red).
The GSG partitioner with CROWN~\cite{zhang2018efficient} propagator ran for 2 sec.
}
\label{fig:tightest_bounds_for_different_conditions}
\end{figure}

\cref{fig:boundary} shows the ability to partition efficiently for different output shapes for a randomly initialized NN with 2 inputs, 2 outputs, and 50 nodes in hidden layer, \emph{i.e.,} $(2,50,2)$, with ReLU activations, and input set $[0,\,1]\times[0,\,1]$.
Each of (a-c) uses GSG with CROWN for 2 seconds.
Recall that SG~\cite{xiang2018output} would only return one output set for (a-c).

For further experiments to quantify the computation-accuracy tradeoff, compare different choices of Partitioners and Propagators, and show how the methods scale to various NN sizes and architectures, interested readers are referred to~\cite{everett2020robustness}.

%% file: nfl_analysis.tex

\section{Analyzing a Neural Feedback Loop}\label{sec:nfl_analysis}

\begin{figure}[t]
	\includegraphics[page=1,width=\linewidth,trim=200 150 200 50,clip]{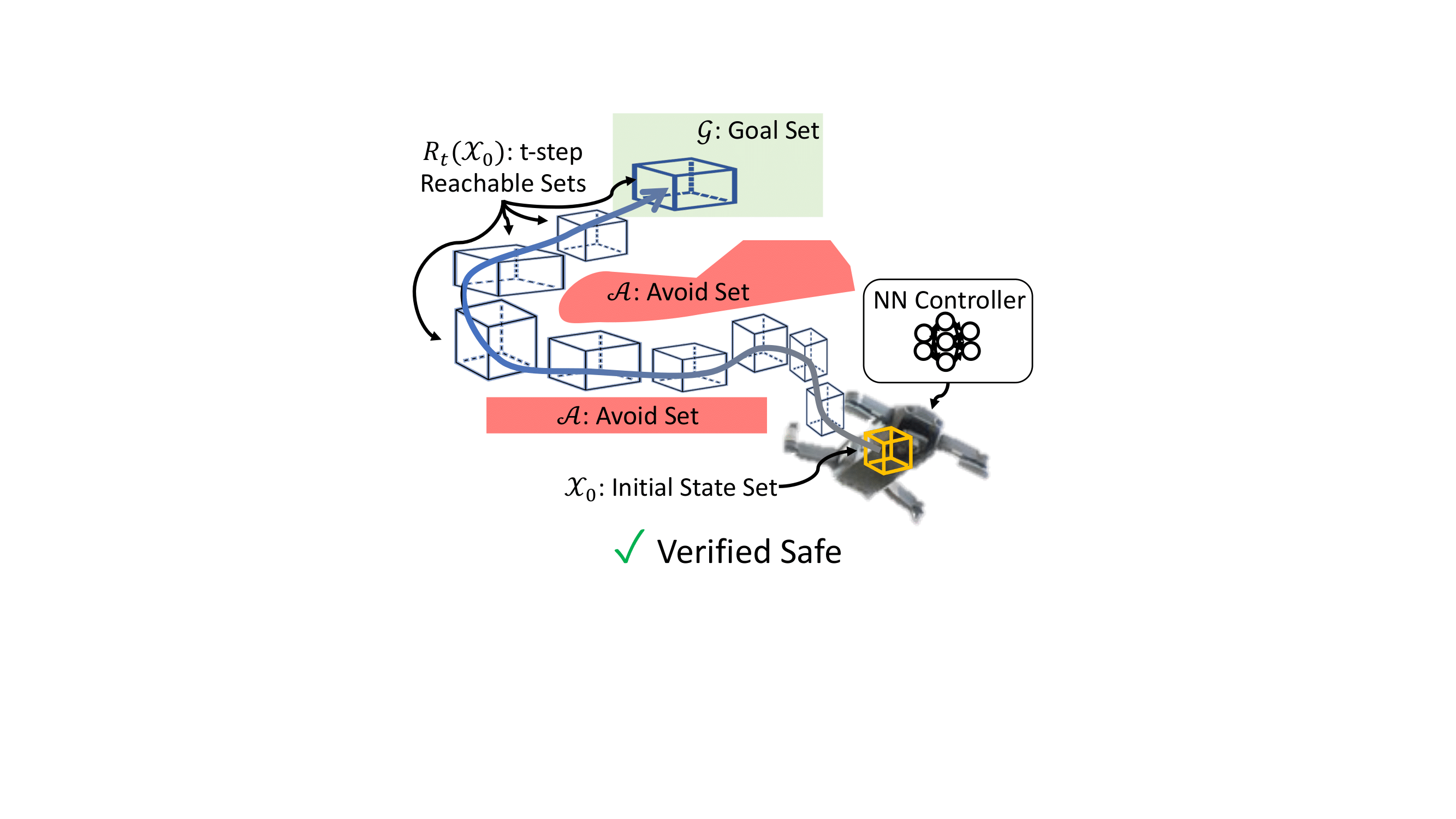}
	\caption{Reachability Analysis. The objective is to compute the blue sets $\mathcal{R}_t(\mathcal{X}_0)$, to ensure a system starting from $\mathcal{X}_0$ (yellow) ends in $\mathcal{G}$ (green) and avoids $\mathcal{A}_0,\mathcal{A}_1$ (red).} 
	\label{fig:problem_cartoon}
\end{figure}

This section of the tutorial develops a framework to \textbf{guarantee that a system with a NN controller will reach the goal states while avoiding undesirable regions of the state space}, as in~\cref{fig:problem_cartoon}.

As noted in~\cref{sec:literature_review}, recent work provides bounds on these reachable sets, but the computationally tractable approaches yield overly conservative bounds (thus cannot be used to verify useful properties), and the methods that yield tighter bounds are too intensive for online computation.
This section bridges the gap by formulating a convex optimization problem for the reachability analysis of closed-loop systems with NN controllers.
While the solutions are less tight than previous (semidefinite program-based) methods, they are substantially faster to compute, and some of those computational time savings can be used to refine the bounds using new input set partitioning techniques, which is shown to dramatically reduce the tightness gap.
The new framework is developed for systems with uncertainty (e.g., measurement and process noise) and nonlinearities (e.g., polynomial dynamics), and thus is shown to be applicable to real-world systems.

\subsection{Preliminaries}\label{sec:preliminaries}

\begin{figure}[t]
    \centering
    \includegraphics[page=4,width=\linewidth,trim=100 100 100 100,clip]{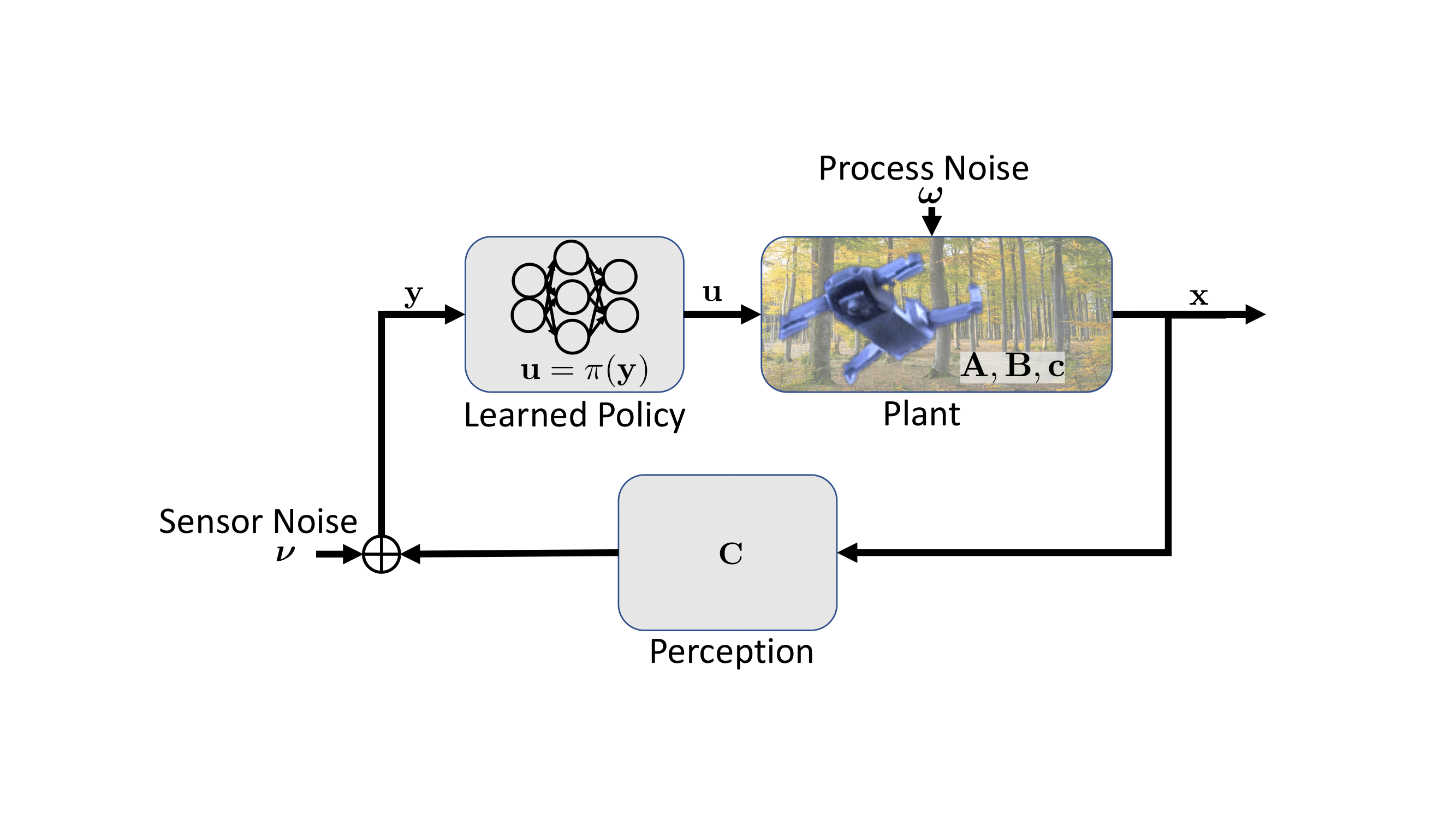}
    \caption{Neural Feedback Loops provide a framework to connect control theoretic ideas with NN analysis and verification.}
    \label{fig:neural_feedback_loop}
\end{figure}

Consider a discrete-time linear time-varying system,
\begin{align}\begin{split}
    \mathbf{x}_{t+1} &= \mathbf{A}_t \mathbf{x}_{t} + \mathbf{B}_t \mathbf{u}_t + \mathbf{c}_t + \bm{\omega}_t = f(\mathbf{x}_{t}; \pi) \\
    \mathbf{y}_{t} &= \mathbf{C}_t^T\mathbf{x}_t + \bm{\nu}_t \\
    \mathbf{u}_t &= \pi(\mathbf{y}_t) \label{eqn:ltv_dynamics} 
\end{split}\end{align}
where $\mathbf{x}_t\in\R^{n_x},\, \mathbf{u}_t\in\R^{n_u},\, \mathbf{y}_t\in\R^{n_y}$ are state, control, and output vectors, $\mathbf{A}_t,\, \mathbf{B}_t,\, \mathbf{C}_t$ are known system matrices, $\mathbf{c}_t\in\R^{n_x}$ is a known exogenous input, and $\bm{\omega}_t\sim\Omega$ and $\bm{\nu}_t\sim N$ are process and measurement noises sampled at each timestep from unknown distributions with known, finite support (i.e., $\bm{\omega}_t\in [\ubar{\bm{\omega}}_t,\, \bar{\bm{\omega}}_t ], \bm{\nu}_t\in [\ubar{\bm{\nu}}_t,\, \bar{\bm{\nu}}_t ]$ element-wise), and output-feedback controller $\pi$ is parameterized by a feed-forward NN with $L$ hidden layers (defined formally in~\cref{sec:nn_analysis:nn_notation}), optionally subject to control constraints, $\mathbf{u}_t\in\mathcal{U}_t$.
If the noise distribution has infinite support (e.g., Gaussian), a confidence interval could be chosen or methods from~\cite{fazlyab2019probabilistic} could be extended to the NFL setting.

This NFL is visualized in~\cref{fig:neural_feedback_loop}.
At each timestep, the state $\mathbf{x}$ enters the perception block and is perturbed by sensor noise $\bm{\nu}$ to create observation $\mathbf{y}$.
The observation is fed into the learned policy, which is a NN that selects the control input $\mathbf{u}$.
The control input enters the plant, which is defined by $\mathbf{A}, \mathbf{B}, \mathbf{c}$ and subject to process noise $\bm{\omega}$.
The output of the plant is the state vector at the next timestep.

We denote $\mathcal{R}_t(\mathcal{X}_0)$ the forward reachable set at time $t$ from a given set of initial conditions $\mathcal{X}_0\subseteq\R^{n_x}$, which is defined by the recursion, $\mathcal{R}_{t+1}(\mathcal{X}_0) = f(\mathcal{R}_t(\mathcal{X}_0); \pi), \mathcal{R}_0(\mathcal{X}_0)=\mathcal{X}_0$.

The finite-time reach-avoid properties verification is defined as follows: Given a goal set $\mathcal{G}\subseteq\R^{n_x}$, a sequence of avoid sets $\mathcal{A}_t\subseteq\R^{n_x}$, and a sequence of reachable set estimates $\mathcal{R}_t\subseteq\R^{n_x}$, determining that every state in the final estimated reachable set will be in the goal set and any state in the estimated reachable sets will not enter an avoid set requires computing set intersections, $\texttt{VERIFIED}(\mathcal{G},\mathcal{A}_{0:N},\mathcal{R}_{0:N})\equiv\mathcal{R}_N\subseteq\mathcal{G}\ \mathrm{\&}\ \mathcal{R}_t\cap\mathcal{A}_t=\emptyset, \forall t\in\{0,\ldots,N\}$.

In the case of our nonlinear closed-loop system~\cref{eqn:ltv_dynamics}, where computing the reachable sets exactly is computationally intractable, we can instead compute outer-approximations of the reachable sets, $\bar{\mathcal{R}}(\mathcal{X}_0)\supseteq\mathcal{R}_t(\mathcal{X}_0)$.
This is useful if the finite-time reach-avoid properties of the system as described by outer-approximations of the reachable sets are verified, because that implies the finite-time reach-avoid properties of the \textit{exact} closed loop system are verified as well.
Furthermore, tight outer-approximations of the reachable sets enable verification of tight goal and avoid set specifications about the exact system, and they reduce the chances of verification being unsuccessful even if the exact system meets the specifications.

\subsection{Relationship with Neural Network Robustness Verification}

A key step in computing reachable sets of the closed-loop system~\cref{eqn:ltv_dynamics} with a NN control policy in a reasonable amount of time is to relax nonlinear constraints induced by the NN's nonlinear activation functions, as described in~\cref{sec:nn_analysis}.

In a closed-loop system, \cref{thm:crown_particular_x} provides a bound on the control output for a \textit{particular} measurement $\mathbf{y}$.
Moreover, if all that is known is $\mathbf{y}\in\mathcal{B}_p(\mathbf{y}_0, \epsilon)$, \cref{thm:crown_particular_x} provides affine relationships between $\mathbf{y}$ and $\mathbf{u}$ (i.e., bounds valid within the known set of possible $\y$).
These relationships enable efficient calculation of bounds on the NN output, using Corollary 3.3 of~\cite{zhang2018efficient}.

We could leverage~\cite{zhang2018efficient} to compute reachable sets by first bounding the possible controls, then bounding the next state set by applying the extreme controls from each state.
This is roughly the approach in~\cite{xiang2020reachable,yang2019efficient}, for example.
However, this introduces excessive conservatism, because both extremes of control would not be applied at every state (barring pathological examples).
To produce tight bounds on the reachable sets, we leverage the relationship between measured output and control in~\cref{sec:nfl_analysis:approach}.

\subsection{Approach}\label{sec:nfl_analysis:approach}

\begin{figure*}[t]
    \centering
    \includegraphics[width=0.8\linewidth,trim=0 150 0 100,clip]{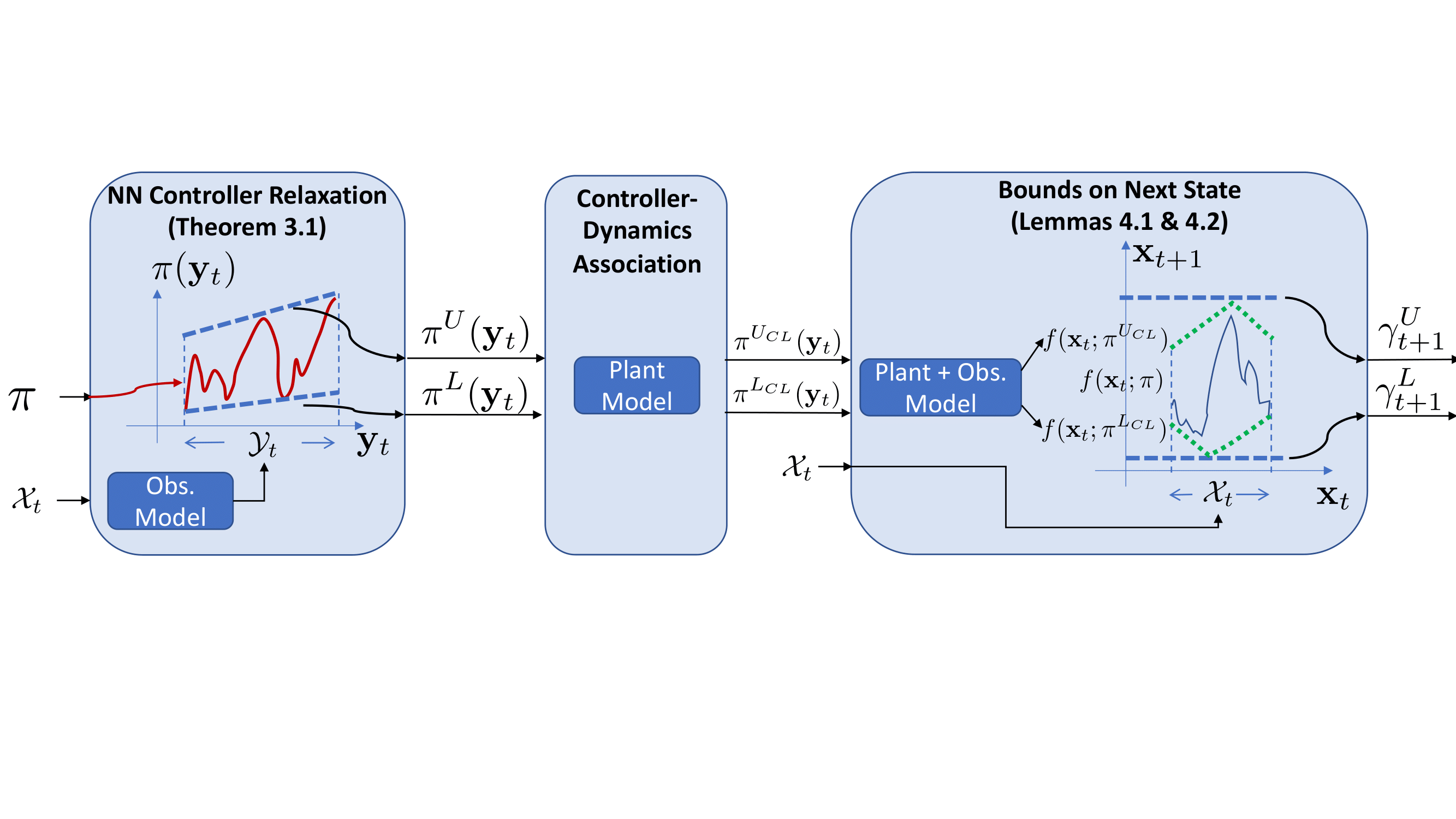}
    \caption{\small Approach Overview for simple 1D system. \cref{thm:crown_particular_x} relaxes the NN to compute the control boundaries $\pi^U, \pi^L$, given the observation $y_t$. After using the system dynamics to associate $\pi^U, \pi^L$ with the next state set, \cref{thm:bounds_on_xt1} computes bounds on the next state, $\gamma^U_{t+1}, \gamma^L_{t+1}$.}
    \label{fig:approach_flow}
\end{figure*}

Recall that our goal is to find the set of all possible next states, $\mathbf{x}_{t+1}\in\Xtt$, given that the current state lies within a known set, $\mathbf{x}_t \in \Xt$.
This will allow us to compute reachable sets recursively starting from an initial set $\Xt=\mathcal{X}_0$.

The approach follows the architecture in~\cref{fig:approach_flow}.
After relaxing the NN controller using~\cref{thm:crown_particular_x}, we associate linearized control bounds with extreme next states (details in Lemma 4.1 of~\cite{everett2021reachability}).
Using the linearized extreme controller, we optimize over all states in the input set to find extreme next states in~\cref{sec:nfl_analysis:approach:bounds_xt1}.
For more detail on how to convert the solutions of the optimization problems into reachable set descriptions, handle control limits, nonlinear terms in the dynamics, and backward reachability, readers are referred to~\cite{everett2021reachability}.

\subsection{Assumptions}
This work assumes that $\Xt$ is described by either:
\begin{itemize}
	\item an $\ell_p$-ball for some norm $p\in[1,\infty]$, radius $\bm{\epsilon$}, and centroid $\xnom$, s.t. $\Xt=\bpxnom$; or
	\item a polytope, for some $\Ain\in\R^{m_\text{in}\times n_x}$, $\mathbf{b}^\text{in}\in\R^{m_\text{in}}$, s.t. $\Xt=\{\mathbf{x}_t\ \lvert\ \Ainbin\}$,
\end{itemize}
and shows how to compute $\Xtt$ as described by either:
\begin{itemize}
	\item an $\ell_{\infty}$-ball with radius $\bm{\epsilon$} and centroid $\xnom$, s.t. $\Xtt=\binfxnom$; or
	\item a polytope for a specified $\Aout\in\mathds{R}^{\mout\times n_x}$, meaning we will compute $\mathbf{b}^\text{out}\in\mathds{R}^{m_{out}}$ s.t. $\Xtt=\{\mathbf{x}_t \in \mathds{R}^{n_x}\ \lvert\ \Aout\mathbf{x}_t\leq \mathbf{b}^\text{out}\}$.
\end{itemize}

We assume that either $\Aout$ is provided (in the case of polytope output bounds), or that $\Aout=\mathbf{I}_{n_x}$ (in the case of $\ell_{\infty}$ output bounds).
Note that we use $\ifacet$ to index polytope facets, $\icontrol$ to index the control vectors, and $\istate$ to index the state vectors.
We assume that $\mathcal{U}_t=\mathds{R}^{n_u}$ (no control input constraints) for cleaner notation, but this assumption is straightforward to relax.

\subsection[Bounds on Next state from initial state set]{Bounds on $\x_{t+1}$ from any $\xinX$}\label{sec:nfl_analysis:approach:bounds_xt1}

To compute $t$-step reachable sets recursively, we form bounds on the next state polytope facet given a \textit{set} of possible current states.
The exact 1-step closed-loop reachability problem is as follows. For each row $\ifacet$ in $\Aout$, solve the following optimization problem,
\begin{alignbox}
\begin{split}
    \left(\boutj\right)^* = \max_{\mathbf{x}_t\in\Xt} &\quad \Aoutj \mathbf{x}_{t+1} \\
    \text{s.t.} &\quad \mathbf{x}_{t+1} = f(\mathbf{x}_t; \pi), \label{eqn:nfl_analysis_exact}
\end{split}
\end{alignbox}
\noindent
where $\left(\bout\right)^*$ defines the tightest description of $\Xtt$ associated with $\Aout$.
However, the nonlinearities in $\pi$ make solving this problem intractable in practice.
Instead, the following relaxation provides bounds on~\cref{eqn:nfl_analysis_exact} (i.e., we compute $\xttubifacet$ s.t. $\Aoutj \mathbf{x}_{t+1} \leq \left(\boutj\right)^* \leq \xttubifacet \ \forall\mathbf{x}_t\in\Xt$).

\begin{lemma}\label[lemma]{thm:bounds_on_xt1}
Given a NN control policy $\pi:\R^{n_y}\to\R^{n_u}$, with $m$ hidden layers, closed-loop dynamics $f: \R^{n_x} \times \Pi \to \R^{n_x}$ as in~\cref{eqn:ltv_dynamics}, and specification matrix $\Aout\in\R^{\mout\times n_x}$, for each $\ifacet\in[\mout-1]$, there exists a fixed value $\xttubifacet$ such that $\forall \xinX$, the inequality $\Aoutj f(\x_t; \pi) \leq \xttubifacet $ holds true, where
\begin{align}
    \xttubifacet &= \max_{\xinX} \Muj \mathbf{x}_t + \Nuj \label{eqn:global_upper_bnd_generic_optimization},
\end{align}
with $\mathbf{M}^U\in\R^{n_x \times n_x}$, $\mathbf{n}^U\in\R^{n_x}$ defined as
\begin{align}
    \Muj &= \left(\Aoutj  \left(\mathbf{A}_{t} + \mathbf{B}_{t} \piclAu_{\ifacet,:,:} \mathbf{C}_t^T \right) \right) \label{eqn:muj_defn}\\
    \Nuj &= \Aoutj \left(\mathbf{B}_{t}\left( \piclAu_{\ifacet,:,:} \Jbar{\AoutjBtPiclAuCt}{\bar{\bm{\nu}}_t}{\ubar{\bm{\nu}}_t} + \piclbu_{:,\ifacet} \right) + \nonumber\right.\\ &\quad\quad\quad\quad\left.\mathbf{c}_{t} + \Jbar{\AoutjBt}{\bar{\bm{\omega}}_t}{\ubar{\bm{\omega}}_t} \right) \label{eqn:nuj_defn},
\end{align}
with each term defined in detail in~\cite{everett2021reachability}, all based on NN weights and dynamics or noise parameters.
Briefly, $\piclAu_{\ifacet,:,:}$ is composed of terms from either $\CROWNAu$ or $\CROWNAl$ (from~\cref{thm:crown_particular_x}) based on the sign of $\Aoutj \mathbf{B}_{t}$ (and $\piclbu_{:,\ifacet}$ is similarly composed of either $\CROWNbu$ or $\CROWNbl$) and $s$ is a selector function similar to \texttt{torch.where}.
Furthermore, lower bounds on the minimzation of~\cref{eqn:nfl_analysis_exact}, $\gamma_{t+1,\mathfrak{j}}^{L}$, can be obtained similarly.
\end{lemma}

The optimization problem in~\cref{eqn:global_upper_bnd_generic_optimization} is a linear program that we solved with \texttt{cvxpy}~\cite{diamond2016cvxpy}.
Note that control limits are straightforward to add to the analysis by adding two ReLUs to the end of the NN~\cite{hu2020reach}.
Further detail is provided in~\cite{everett2021reachability} about converting the state constraints from~\cref{eqn:global_upper_bnd_generic_optimization} into reachable sets and computing tighter reachable sets via parititioning the input set.

Rather than employing an LP solver, the optimization problem in~\cref{eqn:global_upper_bnd_generic_optimization} can be solved in closed-form when $\mathcal{X}_\text{in}$ is described by an $\ell_\infty$-ball.

\begin{lemma}\label[lemma]{thm:closed_form}
In the special case of~\cref{thm:bounds_on_xt1} where $\Xt = \bpxtnom$ for some $p\in[1,\infty)$, the following closed-form expressions are equivalent to~\cref{eqn:global_upper_bnd_generic_optimization},
\begin{align}
    \xttubistate &= \lvert\lvert\bm{\epsilon}\odot \Muj \rvert\rvert_q + \Muj \xtnom + \Nuj \label{eqn:closed_form_upper_bnd}
\end{align}
where ${1/p + 1/q = 1}$ (e.g., $p=\infty, q=1$).
The proof follows from \cite{everett2020certified} and \cite{Weng_2018}.
\end{lemma}

\section{Tighter Reachable Sets by Partitioning the Initial State Set}

Recall that~\cref{sec:nn_analysis} introduced an architecture composed of a \textit{partitioner} and \textit{propagator} for analyzing NNs in isolation.
Here, we extend that framework to closed-loop systems, as visualized in~\cref{fig:system_architecture}.
In particular, CL-CROWN and Reach-SDP represent \textit{closed-loop propagators} (given an initial state set, they compute reachable sets for a trained NN control policy and closed-loop dynamics), and in this section we extend partitioners discussed in~\cite{everett2020robustness} to be \textit{closed-loop partitioners}.
Altogether, we call the nested architecture a \textit{closed-loop analyzer}.

\begin{figure}[t]
    \includegraphics[page=5,width=\linewidth,trim=10 120 0 70,clip]{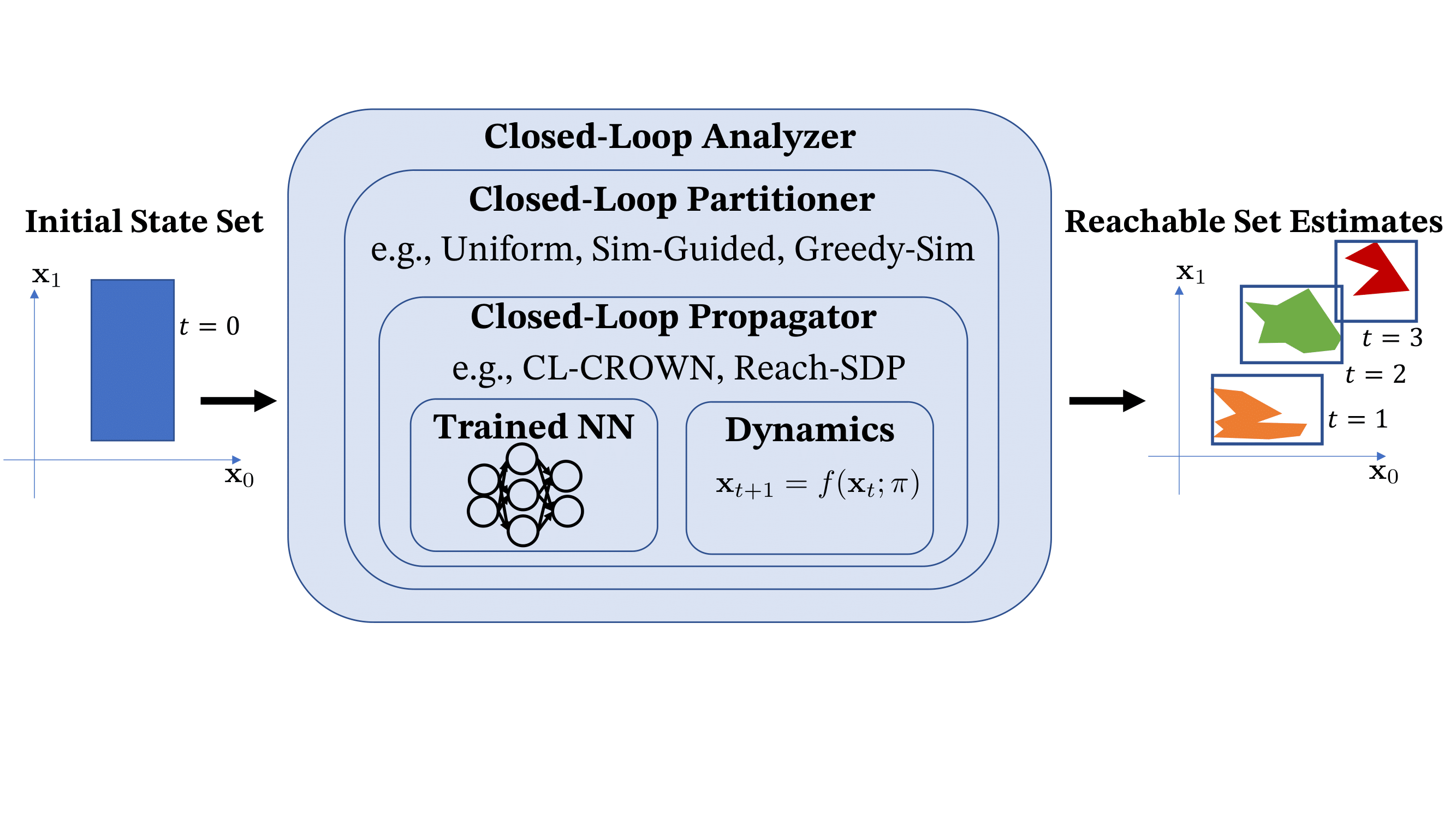}
    \caption{System Architecture. A closed-loop propagator uses the trained NN control policy and dynamics to estimate reachable sets, and a closed-loop partitioner decides how to split the initial state set into pieces. This is the closed-loop extension of the architecture from our prior work~\cite{everett2020robustness}.}
    \label{fig:system_architecture}
\end{figure}

\subsection{Closed-Loop Partitioners}

The simplest closed-loop partitioner splits the input set into a uniform grid.
After splitting $\mathcal{X}_0$ into $\Pi_{\istate=0}^{n_x}\mathbf{r}_\istate$ cells, each cell of $\mathcal{X}_0$ is passed through a closed-loop propagator, CLProp (e.g., CL-CROWN~\cite{everett2021reachability} or Reach-SDP~\cite{hu2020reach}).
For each timestep, the returned estimate of the reachable set for all of $\mathcal{X}_0$ is thus the union of all reachable sets for cells of the $\mathcal{X}_0$ partition.
A closed-loop variant of the greedy simulation-guided partitioner from~\cref{sec:nn_analysis} is described in~\cite{everett2021reachability}.

\subsection{Numerical Experiments}

\begin{table}[t]
\centering
\small{
		\begin{tabular}{|c|c|c|}
			\hline
			 Algorithm                    & Runtime [s]         &   Error \\
			\hline
             \textcolor{blue}{CL-CROWN}~\cite{everett2021reachability}                            & $0.023 \pm 0.000$ &  654556 \\
             CL-SG-IBP~\cite{xiang2020reachable} & $0.888 \pm 0.011$ &   70218 \\
            \hline
			 Reach-SDP~\cite{hu2020reach} & $42.57 \pm 0.54$  &     207 \\
			 \textcolor{blue}{Reach-SDP-Partition}~\cite{everett2021reachability}          & $670.26 \pm 2.91$ &      12 \\
			 \textcolor{blue}{Reach-LP}~\cite{everett2021reachability}                     & $0.017 \pm 0.000$   &    1590 \\
			 \textcolor{blue}{Reach-LP-Partition}~\cite{everett2021reachability}           & $0.263 \pm 0.001$   &      34 \\
			\hline
	    \end{tabular}
	    }
\caption{Reachable Sets for Double Integrator. The first two methods analyze the NN and dynamics separately, leading to high error accumulation, while the next 4 methods analyze the NN and dynamics together. Reach-LP is $4,000\times$ faster to compute but $7\times$ looser than Reach-SDP~\cite{hu2020reach}. Reach-LP-Partition refines the Reach-LP bounds by splitting the input set into 16 subsets, giving $150\times$ faster computation time and $5\times$ tighter bounds than Reach-SDP~\cite{hu2020reach}.}
\label{tab:double_integrator_reachable_set}
\end{table}

\begin{figure}[t]
	\centering
	\begin{subfigure}{0.5\linewidth}
		\centering
		\includegraphics[width=\linewidth]{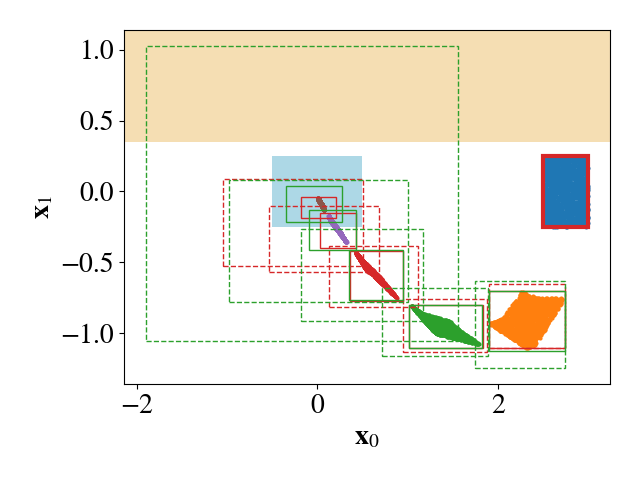}
		\captionsetup{justification=centering}
		\caption{Reachable Set Estimates}
		\label{fig:double_integrator_reachable_set_trajectory}
	\end{subfigure}%
	\begin{subfigure}{0.5\linewidth}
	    \includegraphics[width=\linewidth]{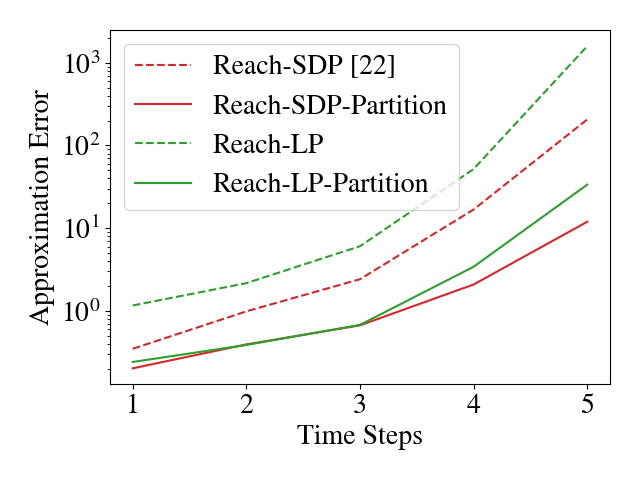}
	    \captionsetup{justification=centering}
	    \caption{Over-approximation error}
		\label{fig:double_integrator_reachable_set_error}
	\end{subfigure}%
	\caption{Reachable Sets for Double Integrator. In (a), all reachable set algorithms bound sampled states across the timesteps, starting from the blue $\mathcal{X}_0$, and the tightness of these bounds is quantified per timestep in (b).}
	\label{fig:double_integrator_reachable_set}
	\vspace{-0.2in}
\end{figure}

Consider a LTI double integrator system from~\cite{hu2020reach} with a NN controller with [5,5] neurons and ReLU activations. 
In~\cref{tab:double_integrator_reachable_set}, 6 algorithms' runtime and error are compared.
Above the solid line, two algorithms are evaluated that separately analyze the NN for control bounds and propagate those bounds through the dynamics, at each timestep.
Below the line, four algorithms are compared, all of which analyze the NN controller and dynamics together, leading to much tighter reachable sets (smaller error). 
These algorithms include Reach-SDP~\cite{hu2020reach}, Reach-LP, and those two algorithms after partitioning the input set into 16 cells.
Reach-LP-Partition provides a $7\times$ improvement in reachable set tightness over the prior state-of-the-art, Reach-SDP~\cite{hu2020reach}, while requiring $4,000\times$ less the computation time. \cref{fig:double_integrator_reachable_set_trajectory} shows sampled trajectories, where each colored cluster of points represents sampled reachable states at a particular timestep (blue$\rightarrow$orange$\rightarrow$green, etc.). This figure also visualizes reachable set bounds for various algorithms. Recall that sampling trajectories could miss possible reachable states, whereas these algorithms are guaranteed to over-approximate the reachable sets, however they provide various degrees of tightness to the sampled points.
We quantitatively compare the rectangle sizes in~\cref{fig:double_integrator_reachable_set_trajectory} by defining the tightness as the ratio of areas between the smallest axis-aligned bounding box on the sampled points and the provided reachable set (minus 1). \cref{fig:double_integrator_reachable_set_error} shows as the system progresses forward in time, all algorithms' tightness get worse, but  Reach-LP-Partition and Reach-SDP-Partition perform the best and similarly.
Additional results in \cite{everett2021reachability} include ablation, runtime analysis, and experiments on nonlinear and high-dimensional systems.

%% file: robust_rl.tex

\newcommand{\carrlfiguredir}{figures/robust_rl}

\section{Robustifying the Implementation of a Learned Policy}\label{sec:robust_rl}

So far, this tutorial has focused on the analysis of a learned policy.
In this section, we describe how the same analysis tools can be used to improve the robustness of a learned policy.
In particular, this section focuses on policies that are trained with deep \acrshort*{rl}.

While there are many techniques for synthesis of \textit{empirically} robust deep RL policies~\cite{mandlekar2017adversarially,Rajeswaran_2017,Muratore_2018,Pinto_2017,morimoto2005robust}, it remains difficult to synthesize a \textit{provably} robust neural network.
Instead, we leverage ideas from~\cref{sec:nn_analysis} that provide a guarantee on how sensitive a trained network's output is to input perturbations for each nominal input~\cite{Ehlers_2017, Katz_2017, Huang_2017b,Lomuscio_2017,Tjeng_2019,Gehr_2018}.
These relaxed methods were previously applied on computer vision or other \acrfull*{sl} tasks.
This section extends the tools for efficient formal neural network robustness analysis (e.g., \cite{Weng_2018,singh2018fast,Wong_2018}) to deep \acrshort*{rl} tasks.
In \acrshort*{rl}, techniques designed for \acrshort*{sl} robustness analysis would simply allow a nominal action to be \textit{flagged} as non-robust if the minimum input perturbation exceeds a robustness threshold (e.g., the system's known level of uncertainty) -- these techniques would not reason about alternative actions.

\textbf{Instead, this section focuses on the \textit{robust decision-making problem}: given a known bound on the input perturbation, what is the best action to take?}
This aligns with the requirement that an agent \textit{must} select an action at each step of an \acrshort*{rl} problem.
The approach uses robust optimization to consider worst-case observational uncertainties and provides certificates on solution quality, making it \textit{certifiably robust}.
The proposed algorithm is called \textbf{C}ertified \textbf{A}dversarial \textbf{R}obustness for Deep \textbf{RL} (\textbf{CARRL}).

\subsection{Preliminaries}

In \acrshort*{rl} problems\footnote{This work considers problems with a continuous state space and discrete action space.}, the state-action value (or, ``Q-value'')
\begin{equation}
Q(\bm{s},a)=\mathop{\mathbb{E}}_{\mathbf{s}'\sim P}\left[\sum_{t=0}^{T}{\gamma^t r(t)}|\mathbf{s}(t{=}0)=\bm{s},\mathrm{a}(t{=}0)=a\right], \nonumber
\end{equation}
expresses the expected accumulation of future reward, $r$, discounted by $\gamma$, received by starting in state $\bm{s}\in\mathbb{R}^n$ and taking one of $d$ discrete actions, $a\in \{a_0, a_1, \ldots, a_{d-1}\}=\mathcal{A}$, with each next state $\mathbf{s}'$ sampled from the (unknown) transition model, $P$.
Furthermore, when we refer to Q-values in this work, we really mean a \acrshort*{dnn} approximation of the Q-values.

Let $\bm{\epsilon}\in\mathbb{R}^n_{\geq 0}$ be the maximum element-wise deviation of the state vector, and let $1\leq p \leq \infty$ parameterize the $\ell_p$-norm.
We define the set of states within this deviation as the $\bm{\epsilon}$-Ball,
\begin{align}
\mathcal{B}_p(\bm{s}_0, \bm{\epsilon}) &= \{\bm{s} : \lim_{\bm{\epsilon}' \to \bm{\epsilon}^+} \lvert\lvert (\bm{s} - \bm{s}_0) \oslash \bm{\epsilon}' \rvert\rvert_p \leq 1\},
\label{eq:eps_ball}
\end{align}
where $\oslash$ denotes element-wise division, and the $\mathrm{lim}$ is only needed to handle the case where $\exists i\ \epsilon_i=0$ (e.g., when the adversary is not allowed to perturb some component of the state, or the agent knows some component of the state vector with zero uncertainty).
The $\ell_p$-norm is defined as ${\lvert\lvert \bm{x} \rvert\rvert_p = (\lvert x_1 \lvert^p + \ldots + \lvert x_n \lvert^p)^{1/p}\ \mathrm{for}\ \bm{x}\in\mathbb{R}^{n}, 1\leq p \leq \infty}$.

\subsection{Certified Robustness in RL vs. SL}

\begin{figure*}[tp]
    \vspace*{-.15in}
    \centering
    \includegraphics[width=\textwidth, clip, trim=0 320 0 0]{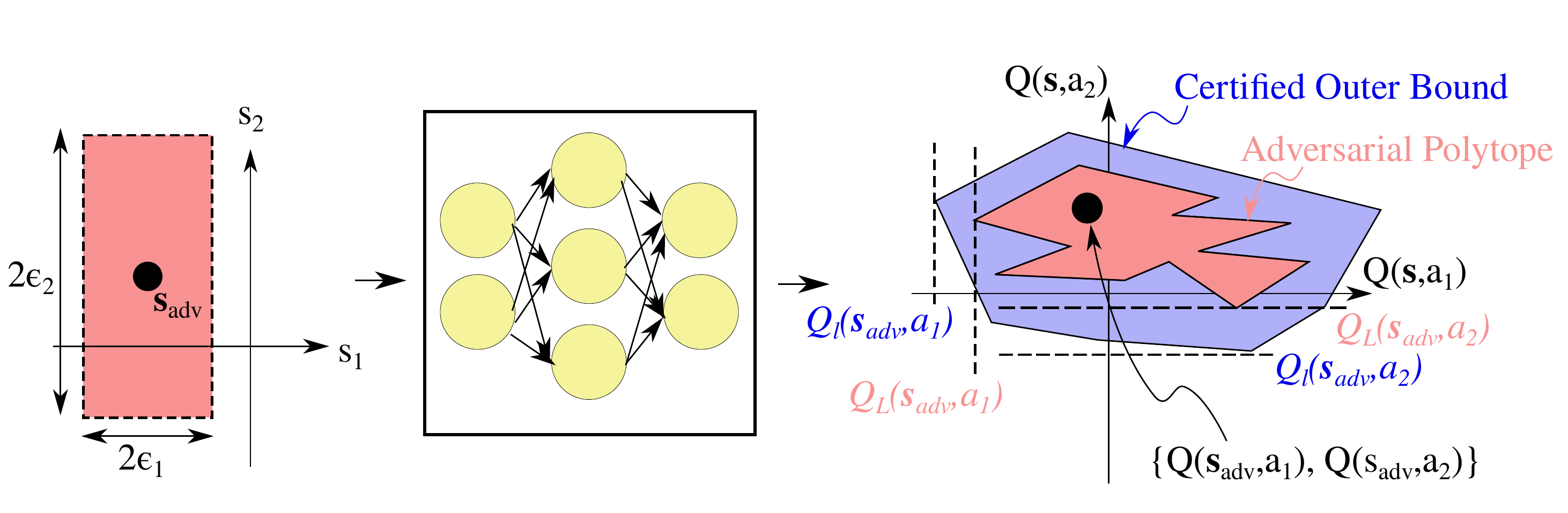}
    \caption[State Uncertainty Propagated Through \acrlong{dqn}]{
    State Uncertainty Propagated Through \acrlong{dqn}.
    The red region (left) represents bounded state uncertainty (\mbox{an $\ell_{\infty}\ \bm{\epsilon}$-ball}) around the observed state, $\bm{s}_\text{adv}$.
    A neural network maps this set of possible inputs to a polytope (red) of possible outputs (Q-values in \acrshort*{rl}).
    This work's extension of~\cite{Weng_2018} provides an outer bound (blue) on that polytope.
    This work then modifies the \acrshort*{rl} action-selection rule by considering lower bounds, $Q_l$, on the blue region for each action.
    In this 2-state, 2-action example, our algorithm would select action 2, since $Q_l(\bm{s}_\text{adv},a_2)>Q_l(\bm{s}_\text{adv},a_1)$, i.e. the worst possible outcome from action 2 is better than the worst possible outcome from action 1, given an $\bm{\epsilon}$-ball uncertainty and a pre-trained DQN.
    }
    \label{fig:state_uncertainty_cartoon}
\end{figure*}

The formal robustness analysis approaches can be visualized in~\cref{fig:state_uncertainty_cartoon} in terms of a 2-state, 2-action deep \acrshort*{rl} problem.
Deep Q-Network (DQN)~\cite{Mnih_2015} is the example deep \acrshort*{rl} algorithm used throughout this section, but similar ideas apply to other value- and policy-based algorithms.
For a given state uncertainty set (red, left), exact analysis methods reason over the exact adversarial polytope (red, right), i.e., the image of the state uncertainty set through the network.
The lower bounds (w.r.t. all states near $\bm{s}_\text{adv}$) of the $Q$-value for each action are shown as the dashed lines labeled $Q_L(\mathbf{s}_\text{adv}, \cdot)$.
By relaxing the problem as described in~\cref{sec:nn_analysis}, bounds on the DQN's reachable set can be obtained, shown as $Q_l(\mathbf{s}_\text{adv}, \cdot)$.

The analysis techniques in~\cref{sec:nn_analysis} are often formulated to solve the \textit{robustness verification problem}: determine whether the calculated set of possible network outputs crosses a decision hyperplane (a line of slope 1 through the origin, in this example) -- if it does cross, the classifier is deemed ``not robust'' to the input uncertainty.
Our approach instead solves the \textit{robust decision-making problem}: determine the best action considering worst-case outcomes, $Q_l(\bm{s}_\text{adv}, a_1), Q_l(\bm{s}_\text{adv}, a_2)$, denoted by the dashed lines in \cref{fig:state_uncertainty_cartoon}.

The inference phase of reinforcement learning (RL) and supervised learning (SL) are similar in that the policy chooses an action/class given some current input. However, there are at least three key differences between the proposed certified robustness for RL and existing certification methods for SL:

\begin{enumerate}
    \item Existing methods for SL certification do not actually change the decision, they simply provide an additional piece of information (``is this decision sensitive to the set of possible inputs? yes/no/unsure'') alongside the nominal decision.
    Most SL certification papers do not discuss what to do if the decision \textit{is} sensitive, which we identify as a key technical gap since in
    RL it is not obvious what to do with a sensitivity flag, as an action still needs to be taken.
    Thus, instead of just returning the nominal decision plus a sensitivity flag, the method in our paper makes a robust decision that considers the full set of inputs.
    \item In SL, there is a ``correct'' output for each input (the true class label).
    In RL, the correct action is less clear because the actions need to be compared/evaluated in terms of their impact on the system.
    Furthermore, in RL, some actions could be much worse than others, whereas in SL typically all ``wrong'' outputs are penalized the same.
    Our method proposes using the Q-value encoded in a DNN as a way of comparing actions under state uncertainty.
    \item In SL, a wrong decision incurs an immediate cost, but that decision does not impact the future inputs/costs.
    In RL, actions affect the system state, which \textit{does} influence future inputs/rewards.
    This paper uses the value function to encode that information about the future, which we believe is an important first step towards considering feedback loops in the certification process.
\end{enumerate}

To summarize, despite SL and RL inference appearing similarly, there are several key differences in the problem contexts that motivate a fundamentally different certification framework for RL.
For a longer discussion of the literature on attacks, defenses, and robustness in RL, see~\cite{everett2020certified,ilahi2021challenges}.

\subsection{Robustness Analysis}

This section aims to find the action that maximizes state-action value under a worst-case perturbation of the observation by sensor noise or an adversary.
The adversary perturbs the true state, $\bm{s}_0$, to another state, ${\bm{s}_\text{adv} \in \mathcal{B}_{p_\text{adv}}(\bm{s}_{0}, \bm{\epsilon}_\text{adv})}$, within the $\bm{\epsilon}_\text{adv}$-ball.
The ego agent only observes the perturbed state, ${\bm{s}_\text{adv}}$.
As displayed in~\cref{fig:state_uncertainty_cartoon}, let the worst-case state-action value, $Q_L$, for a given action, $a_j$, be 
\begin{align}
Q_L(\bm{s}_\text{adv},a_j) &= \min_{\bm{s} \in \mathcal{B}_{p_\text{adv}}(\bm{s}_\text{adv}, \bm{\epsilon}_\text{adv})} Q(\bm{s}, a_j), \label{eq:qlj_lower_bnd}
\end{align}
for all states $\bm{s}$ inside the $\bm{\epsilon}_\text{adv}$-Ball around the observation, $\bm{s}_\text{adv}$.

As introduced in~\cref{sec:nn_analysis}, the goal of the analysis is to compute a guaranteed lower bound, $Q_l(\bm{s},a_j)$, on the minimum state-action value, that is, $Q_l(\bm{s},a_j) \leq Q_L(\bm{s}, a_j)$.
The key idea is to pass interval bounds $[\bm{l}^{(0)}, \bm{u}^{(0)}] = [\bm{s}_\text{adv} - \bm{\epsilon}_\text{adv}, \bm{s}_\text{adv} + \bm{\epsilon}_\text{adv}]$ from the \acrshort*{dnn}'s input layer to the output layer.



\subsection{Problem Setup} \label{sec:robust_rl:approach:problem_setup}

\begin{figure*}
	\centering\includegraphics[page=1,trim=0 150 90 150, clip, width=1.0\textwidth, angle = 0]{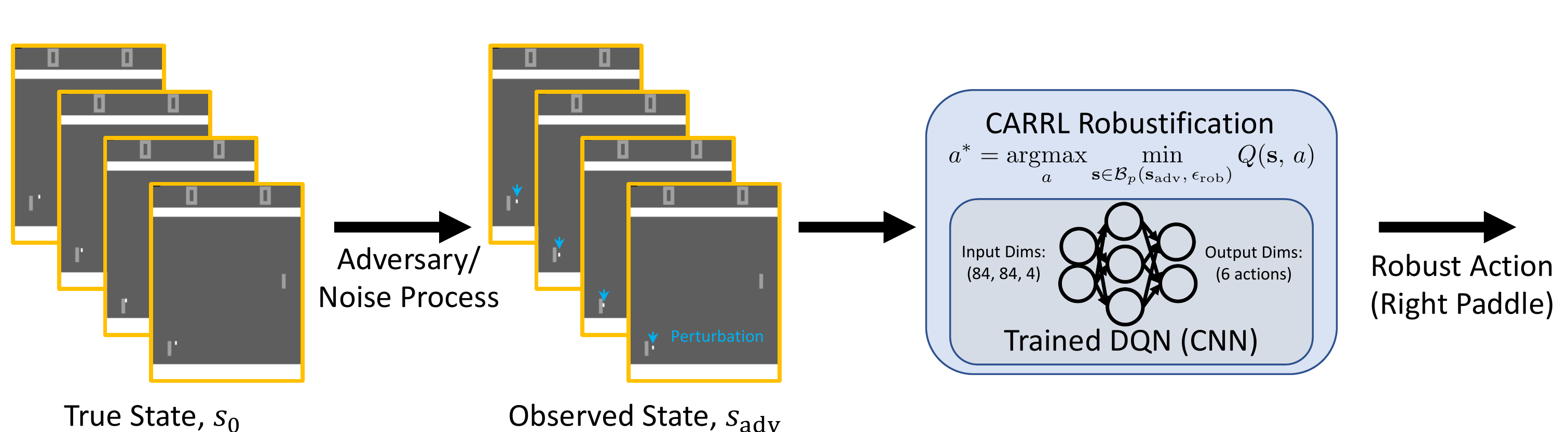}
\caption[Adversary \& Defense in Atari Pong.]{
Adversary \& Defense in Atari Pong.
At each timestep, given the true state (the past 4 $210 \times 160$ grayscale image frames), the adversary finds the pixels corresponding to the ball and moves them downward by $\epsilon_\text{adv}$ pixels.
The CARRL algorithm enumerates the possible un-perturbed states, according to $\epsilon_\text{rob}$, re-shapes the observation to $(84, 84, 4)$, and makes a DQN forward pass to compute $Q_L$ from~\cref{eq:robust_optimal_action_rule}.
The resulting robust-optimal action is implemented, which moves the environment to a new true state, and the cycle repeats.
}
\label{fig:carrl_pong_arch}
\end{figure*}

In an offline training phase, an agent uses a deep \acrshort*{rl} algorithm, here \acrshort*{dqn}~\citep{Mnih_2015}, to train a \acrshort*{dnn} that maps non-corrupted state observations, $\bm{s}_0$, to state-action values, $Q(\bm{s}_0,a)$.
Action selection during training uses the nominal cost function, $a^*_\text{nom} = \argmax_{a_j} Q(\bm{s}_0, a_j)$.

During online execution, the agent only receives corrupted state observations from the environment.
The robustness analysis node uses the \acrshort*{dnn} architecture, \acrshort*{dnn} weights, and robustness hyperparameters, $\bm{\epsilon}_\text{rob}, p_\text{rob}$, to compute lower bounds on possible Q-values for robust action selection.

\subsection{Optimal cost function under worst-case perturbation} \label{sec:robust_rl:approach:optimal_cost_function}
We assume that the training process causes the network to converge to the optimal value function, $Q^*(\bm{s}_0,a)$ and focus on the challenge of handling perturbed observations during execution.
Thus, we consider robustness to an adversary that perturbs the true state, $\bm{s}_0$, within a small perturbation, $\bm{\epsilon}_\text{adv}$, into the worst-possible state observation, $\bm{s}_\text{adv}$.
The adversary assumes that the \acrshort*{rl} agent follows a nominal policy (as in, e.g., \acrshort*{dqn}) of selecting the action with highest Q-value at the current observation.
A worst possible state observation, $\bm{s}_\text{adv}$, is therefore any one which causes the \acrshort*{rl} agent to take the action with lowest Q-value in the true state, $\bm{s}_0$:
\begin{align}
\bm{s}_\text{adv} \in \{\bm{s}:\; &\bm{s} \in \mathcal{B}_{p_\text{adv}}(\bm{s}_0, \bm{\epsilon}_\text{adv})\, \text{and}\,\nonumber \\&
\argmax_{a_j} Q(\bm{s}, a_j) = \argmin_{a_j} Q(\bm{s}_0, a_j)\}.
\label{eq:s_adv}
\end{align}
This set could be computationally intensive to compute and/or empty -- an approximation is described in~\cite{everett2020certified}.

After the agent receives the state observation picked by the adversary, the agent selects an action.
Instead of trusting the observation (and thus choosing the worst action for the true state), the agent could leverage the fact that the true state, $\bm{s}_0$, must be somewhere inside an $\bm{\epsilon}_\text{adv}$-Ball around $\bm{s}_\text{adv}$ (i.e., ${\bm{s}_0 \in \mathcal{B}_{p_\text{adv}}(\bm{s}_\text{adv}, \bm{\epsilon}_\text{adv}))}$.

For the sake of the tutorial, assume that the adversary and robust agent's parameters are the same (i.e., ${\bm{\epsilon}_\text{rob}=\bm{\epsilon}_\text{adv}}$ and ${p_\text{rob}=p_\text{adv}}$ to ensure ${\mathcal{B}_{p_\text{rob}}(\bm{s}_\text{adv}, \bm{\epsilon}_\text{rob}))=\mathcal{B}_{p_\text{adv}}(\bm{s}_\text{adv}, \bm{\epsilon}_\text{adv}))}$).
Empirical effects of tuning ${\bm{\epsilon}_\text{rob}}$ to other values are explored in~\cite{everett2020certified}.

The agent evaluates each action by calculating the worst-case Q-value under all possible true states.
\begin{definition}
In accordance with the robust decision making problem, the robust-optimal action, $a^*$, is defined here as one with the highest Q-value under the worst-case perturbation,
\begin{alignbox}
 a^* &= \argmax_{a_j}\underbrace{\min_{\bm{s}\in \mathcal{B}_{p_\text{rob}}(\bm{s}_\text{adv}, \bm{\epsilon}_\text{rob})} Q(\bm{s},a_j)}_{Q_L(\bm{s}_\text{adv}, a_j)}.\label{eq:robust_optimal_action_rule}
\end{alignbox}
\end{definition}
As described above, computing $Q_L(\bm{s}_\text{adv},a_j)$ exactly is too computationally intensive for real-time decision-making.

\begin{definition}
In \acrshort*{carrl}, the action, $a_\text{CARRL}$, is selected by approximating $Q_L(\bm{s}_\text{adv},a_j)$ with $Q_l(\bm{s}_\text{adv}, a_j)$, its guaranteed lower bound across all possible states $\bm{s}\in\mathcal{B}_{p_\text{rob}}(\bm{s}_\text{adv},\bm{\epsilon}_\text{rob})$, so that:
\begin{equation}
 a_\text{CARRL} = \argmax_{a_j} Q_l(\bm{s}_\text{adv}, a_j).
\label{eq:opt_cost_fn} 
\end{equation}
\end{definition}
Conditions for optimality (${a^*=a_\text{CARRL}}$), a certificate of sub-optimality, and a closed-form solution for \cref{eq:opt_cost_fn} with a vector-$\mathbf{\epsilon}$-ball are provided in~\cite{everett2020certified}.

\subsection{Other Directions}
\cite{everett2020certified} describes other uses of the robustness analysis tools from~\cref{sec:nn_analysis}, including
\begin{itemize}
	\item for policy-based RL algorithms (e.g., SAC~\cite{haarnoja2018soft,christodoulou2019soft})
	\item probabilistic robustness by appropriate choice of ${\bm{\epsilon}_\text{rob}}$ (e.g., for a Gaussian sensor model with known standard deviation of measurement error, $\bm{\sigma}_{sensor}$, one could set ${\bm{\epsilon}_\text{rob}=2\bm{\sigma}_{sensor}}$ to yield actions that account for the worst-case outcome with 95\% confidence.)
	\item a different robustness paradigm, which prefers actions with low sensitivity to changes in the input while potentially sacrificing performance
\end{itemize}

\subsection{Results}

\begin{figure*}[t]
      \begin{subfigure}{0.5\linewidth}
      \centering
        \includegraphics[width=0.7\textwidth, trim=20 0 40 10, clip]{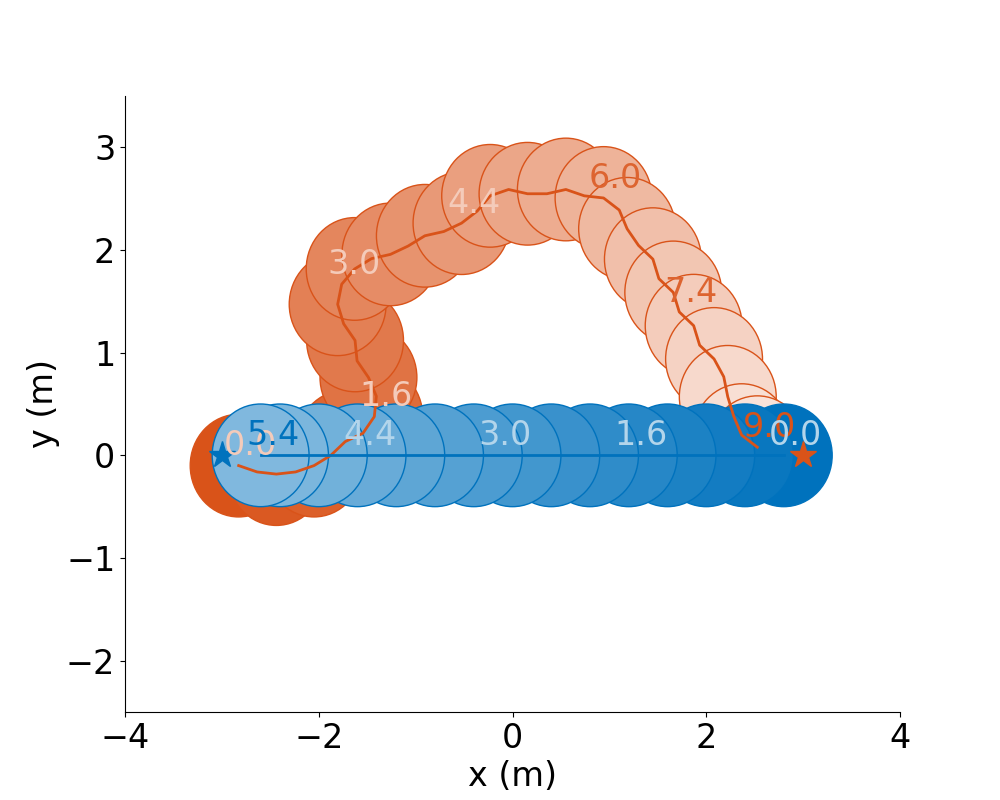}%
                        \captionsetup{justification=centering}
          \caption{CROWN (highly conservative)} 
        \label{fig:rl_crown}
      \end{subfigure}%
      \begin{subfigure}{0.5\linewidth}
            \centering
        \includegraphics[width=0.7\textwidth, trim=20 0 40 10, clip]{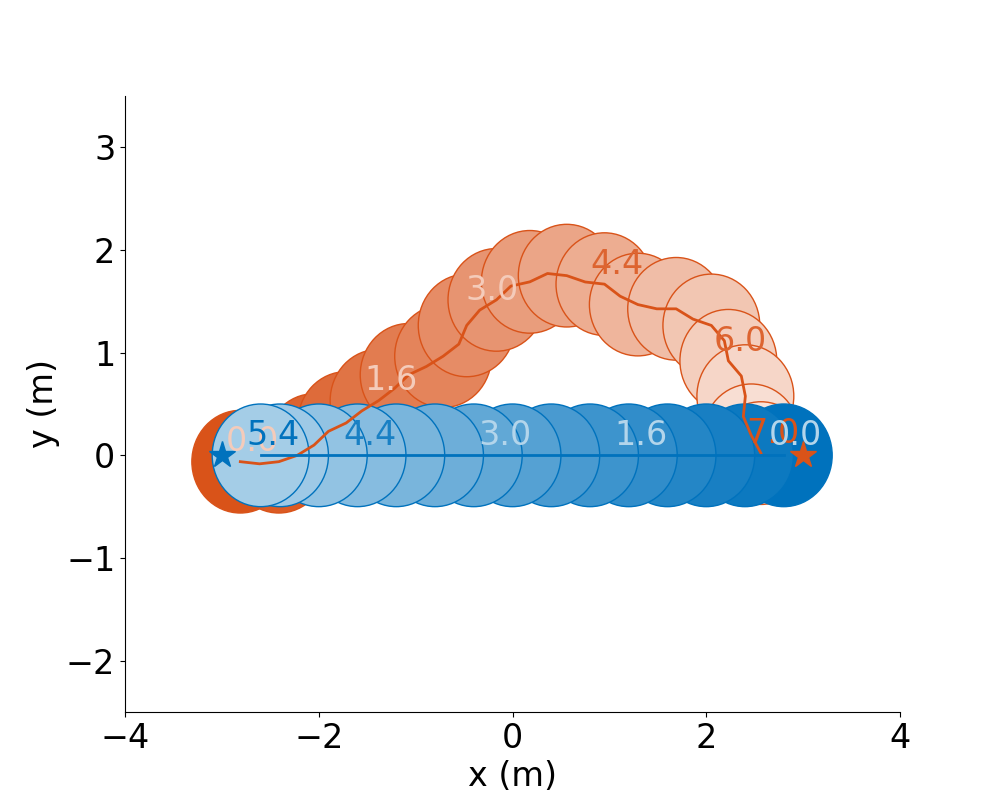}%
                \captionsetup{justification=centering}
        \caption{GSG-CROWN (less conservative)}
        \label{fig:rl_gsgcrown}
      \end{subfigure}
  \caption{
Multiagent collision avoidance under uncertainty.
In (a), a robust but conservative trajectory from a robust RL formulation~\cite{everett2020certified} that used CROWN/Fast-Lin~\cite{zhang2018efficient} to estimate worst-case Q-values under uncertainty on obstacle positions.
In (b), the proposed GSG-CROWN algorithm enables the orange agent to reach its goal faster (7 vs. 9 sec) while still avoiding the blue agent.
This improved behavior is a result of tighter estimates of worst-case Q-values.}
\label{fig:applications}
\vspace{-0.1in}
\end{figure*}

In~\cite{everett2020certified}, CARRL is deployed for Pong (\cref{fig:carrl_pong_arch}), multiagent collision avoidance, and the cartpole task.
For each task, the algorithm is shown to provide improvements in performance compared to vanilla DQN for various noise or adversary strengths.
In addition, bringing partitioning ideas from~\cref{sec:nn_analysis} into robust RL can provide tighter bounds, $Q_l$, which can lead to better closed-loop performance.
\cref{fig:applications} shows an example of this for multiagent collision avoidance under uncertainty.

%% file: main.bbl
\begin{thebibliography}{100}

\bibitem{lamnabhi2017systems}
F.~Lamnabhi-Lagarrigue, A.~Annaswamy, S.~Engell, A.~Isaksson, P.~Khargonekar,
  R.~M. Murray, H.~Nijmeijer, T.~Samad, D.~Tilbury, and P.~Van~den Hof,
  ``Systems \& control for the future of humanity, research agenda: Current and
  future roles, impact and grand challenges,'' {\em Annual Reviews in Control},
  vol.~43, pp.~1--64, 2017.

\bibitem{Szegedy_2014}
C.~Szegedy, W.~Zaremba, I.~Sutskever, J.~Bruna, D.~Erhan, I.~Goodfellow, and
  R.~Fergus, ``Intriguing properties of neural networks,'' in {\em
  International Conference on Learning Representations (ICLR)}, 2014.

\bibitem{Huang_2017}
S.~Huang, N.~Papernot, I.~Goodfellow, Y.~Duan, and P.~Abbeel, ``Adversarial
  attacks on neural network policies,'' 2017.

\bibitem{behzadan2017vulnerability}
V.~Behzadan and A.~Munir, ``Vulnerability of deep reinforcement learning to
  policy induction attacks,'' in {\em International Conference on Machine
  Learning and Data Mining in Pattern Recognition (MLDM)}, pp.~262--275,
  Springer, 2017.

\bibitem{yang2020enhanced}
C.-H.~H. Yang, J.~Qi, P.-Y. Chen, Y.~Ouyang, I.-T.~D. Hung, C.-H. Lee, and
  X.~Ma, ``Enhanced adversarial strategically-timed attacks against deep
  reinforcement learning,'' in {\em ICASSP 2020-2020 IEEE International
  Conference on Acoustics, Speech and Signal Processing (ICASSP)},
  pp.~3407--3411, IEEE, 2020.

\bibitem{mandlekar2017adversarially}
A.~Mandlekar, Y.~Zhu, A.~Garg, L.~Fei-Fei, and S.~Savarese, ``Adversarially
  robust policy learning: Active construction of physically-plausible
  perturbations,'' in {\em 2017 IEEE/RSJ International Conference on
  Intelligent Robots and Systems (IROS)}, pp.~3932--3939, IEEE, 2017.

\bibitem{Rajeswaran_2017}
A.~Rajeswaran, S.~Ghotra, B.~Ravindran, and S.~Levine, ``Epopt: Learning robust
  neural network policies using model ensembles,'' in {\em International
  Conference on Learning Representations (ICLR)}, 2017.

\bibitem{Muratore_2018}
F.~Muratore, F.~Treede, M.~Gienger, and J.~Peters, ``Domain randomization for
  simulation-based policy optimization with transferability assessment,'' in
  {\em 2nd Annual Conference on Robot Learning, CoRL 2018, Z{\"{u}}rich,
  Switzerland, 29-31 October 2018, Proceedings}, pp.~700--713, 2018.

\bibitem{Pinto_2017}
L.~Pinto, J.~Davidson, R.~Sukthankar, and A.~Gupta, ``Robust adversarial
  reinforcement learning,'' in {\em Proceedings of the 34th International
  Conference on Machine Learning (ICML)} (D.~Precup and Y.~W. Teh, eds.),
  vol.~70 of {\em Proceedings of Machine Learning Research}, (International
  Convention Centre, Sydney, Australia), pp.~2817--2826, PMLR, 06--11 Aug 2017.

\bibitem{morimoto2005robust}
J.~Morimoto and K.~Doya, ``Robust reinforcement learning,'' {\em Neural
  computation}, vol.~17, no.~2, pp.~335--359, 2005.

\bibitem{gleave2019adversarial}
A.~Gleave, M.~Dennis, N.~Kant, C.~Wild, S.~Levine, and S.~Russell,
  ``Adversarial policies: Attacking deep reinforcement learning,'' 2020.

\bibitem{Uther_1997}
W.~Uther and M.~Veloso, ``Adversarial reinforcement learning,'' tech. rep., In
  Proceedings of the AAAI Fall Symposium on Model Directed Autonomous Systems,
  1997.

\bibitem{everett2020robustness}
M.~Everett, G.~Habibi, and J.~P. How, ``Robustness analysis of neural networks
  via efficient partitioning with applications in control systems,'' {\em IEEE
  Control Systems Letters}, 2020.

\bibitem{everett2020certified}
M.~Everett, B.~Lutjens, and J.~P. How, ``Certifiable robustness to adversarial
  state uncertainty in deep reinforcement learning,'' {\em IEEE Transactions on
  Neural Networks and Learning Systems}, pp.~1--15, 2021.

\bibitem{everett2021efficient}
M.~Everett, G.~Habibi, and J.~P. How, ``Efficient reachability analysis of
  closed-loop systems with neural network controllers,'' in {\em IEEE
  International Conference on Robotics and Automation (ICRA)}, 2021.

\bibitem{everett2021reachability}
M.~Everett, G.~Habibi, C.~Sun, and J.~P. How, ``Reachability analysis of neural
  feedback loops,'' {\em arXiv preprint arXiv:2108.04140}, 2021.

\bibitem{widrow1964pattern}
B.~Widrow, ``Pattern-recognizing control systems,'' {\em Compurter and
  Information Sciences}, 1964.

\bibitem{fu1970learning}
K.-S. Fu, ``Learning control systems--review and outlook,'' {\em IEEE
  transactions on Automatic Control}, vol.~15, no.~2, pp.~210--221, 1970.

\bibitem{psaltis1988multilayered}
D.~Psaltis, A.~Sideris, and A.~A. Yamamura, ``A multilayered neural network
  controller,'' {\em IEEE control systems magazine}, vol.~8, no.~2, pp.~17--21,
  1988.

\bibitem{li1989neural}
W.~Li and J.-J.~E. Slotine, ``Neural network control of unknown nonlinear
  systems,'' in {\em 1989 American Control Conference}, pp.~1136--1141, IEEE,
  1989.

\bibitem{narendra1991adaptive}
K.~S. Narendra, ``Adaptive control using neural networks,'' {\em Neural
  networks for control}, vol.~3, 1991.

\bibitem{fukushima1982neocognitron}
K.~Fukushima and S.~Miyake, ``Neocognitron: A self-organizing neural network
  model for a mechanism of visual pattern recognition,'' in {\em Competition
  and cooperation in neural nets}, pp.~267--285, Springer, 1982.

\bibitem{hochreiter1997long}
S.~Hochreiter and J.~Schmidhuber, ``Long short-term memory,'' {\em Neural
  computation}, vol.~9, no.~8, pp.~1735--1780, 1997.

\bibitem{sanders2010cuda}
J.~Sanders and E.~Kandrot, {\em CUDA by example: an introduction to
  general-purpose GPU programming}.
\newblock Addison-Wesley Professional, 2010.

\bibitem{jouppi2017datacenter}
N.~P. Jouppi, C.~Young, N.~Patil, D.~Patterson, G.~Agrawal, R.~Bajwa, S.~Bates,
  S.~Bhatia, N.~Boden, A.~Borchers, {\em et~al.}, ``In-datacenter performance
  analysis of a tensor processing unit,'' in {\em Proceedings of the 44th
  annual international symposium on computer architecture}, pp.~1--12, 2017.

\bibitem{jia2014caffe}
Y.~Jia, E.~Shelhamer, J.~Donahue, S.~Karayev, J.~Long, R.~Girshick,
  S.~Guadarrama, and T.~Darrell, ``Caffe: Convolutional architecture for fast
  feature embedding,'' in {\em Proceedings of the 22nd ACM international
  conference on Multimedia}, pp.~675--678, 2014.

\bibitem{abadi2016tensorflow}
M.~Abadi, P.~Barham, J.~Chen, Z.~Chen, A.~Davis, J.~Dean, M.~Devin,
  S.~Ghemawat, G.~Irving, M.~Isard, {\em et~al.}, ``Tensorflow: A system for
  large-scale machine learning,'' in {\em 12th $\{$USENIX$\}$ symposium on
  operating systems design and implementation ($\{$OSDI$\}$ 16)}, pp.~265--283,
  2016.

\bibitem{paszke2019pytorch}
A.~Paszke, S.~Gross, F.~Massa, A.~Lerer, J.~Bradbury, G.~Chanan, T.~Killeen,
  Z.~Lin, N.~Gimelshein, L.~Antiga, {\em et~al.}, ``Pytorch: An imperative
  style, high-performance deep learning library,'' {\em arXiv preprint
  arXiv:1912.01703}, 2019.

\bibitem{deng2009imagenet}
J.~Deng, W.~Dong, R.~Socher, L.-J. Li, K.~Li, and L.~Fei-Fei, ``Imagenet: A
  large-scale hierarchical image database,'' in {\em 2009 IEEE conference on
  computer vision and pattern recognition}, pp.~248--255, Ieee, 2009.

\bibitem{bellemare13arcade}
M.~G. {Bellemare}, Y.~{Naddaf}, J.~{Veness}, and M.~{Bowling}, ``The arcade
  learning environment: An evaluation platform for general agents,'' {\em
  Journal of Artificial Intelligence Research}, vol.~47, pp.~253--279, jun
  2013.

\bibitem{mott1995stella}
B.~W. Mott, T.~Team, {\em et~al.}, ``Stella: a multiplatform atari 2600 vcs
  emulator,'' 1995.

\bibitem{Brockman_2016}
G.~Brockman, V.~Cheung, L.~Pettersson, J.~Schneider, J.~Schulman, J.~Tang, and
  W.~Zaremba, ``Openai gym,'' 2016.

\bibitem{srivastava2014dropout}
N.~Srivastava, G.~Hinton, A.~Krizhevsky, I.~Sutskever, and R.~Salakhutdinov,
  ``Dropout: a simple way to prevent neural networks from overfitting,'' {\em
  The journal of machine learning research}, vol.~15, no.~1, pp.~1929--1958,
  2014.

\bibitem{goodfellow2014generative}
I.~J. Goodfellow, J.~Pouget-Abadie, M.~Mirza, B.~Xu, D.~Warde-Farley, S.~Ozair,
  A.~C. Courville, and Y.~Bengio, ``Generative adversarial nets,'' in {\em
  NIPS}, 2014.

\bibitem{tschannen2018autoencoder}
M.~Tschannen, O.~F. Bachem, and M.~Lučić, ``Recent advances in
  autoencoder-based representation learning,'' in {\em Bayesian Deep Learning
  Workshop, NeurIPS}, 2018.

\bibitem{howard2017mobilenets}
A.~G. Howard, M.~Zhu, B.~Chen, D.~Kalenichenko, W.~Wang, T.~Weyand,
  M.~Andreetto, and H.~Adam, ``Mobilenets: Efficient convolutional neural
  networks for mobile vision applications,'' {\em arXiv preprint
  arXiv:1704.04861}, 2017.

\bibitem{pomerleau1989alvinn}
D.~A. Pomerleau, ``Alvinn: An autonomous land vehicle in a neural network,''
  tech. rep., CARNEGIE-MELLON UNIV PITTSBURGH PA ARTIFICIAL INTELLIGENCE AND
  PSYCHOLOGY~…, 1989.

\bibitem{levine2014learning}
S.~Levine and P.~Abbeel, ``Learning neural network policies with guided policy
  search under unknown dynamics.,'' in {\em NIPS}, vol.~27, pp.~1071--1079,
  Citeseer, 2014.

\bibitem{tan2018sim}
J.~Tan, T.~Zhang, E.~Coumans, A.~Iscen, Y.~Bai, D.~Hafner, S.~Bohez, and
  V.~Vanhoucke, ``Sim-to-real: Learning agile locomotion for quadruped
  robots.,'' in {\em Robotics: Science and Systems}, 2018.

\bibitem{bansal2017hamilton}
S.~Bansal, M.~Chen, S.~Herbert, and C.~J. Tomlin, ``Hamilton-jacobi
  reachability: A brief overview and recent advances,'' in {\em 2017 IEEE 56th
  Annual Conference on Decision and Control (CDC)}, pp.~2242--2253, IEEE, 2017.

\bibitem{lutter2018deep}
M.~Lutter, C.~Ritter, and J.~Peters, ``Deep lagrangian networks: Using physics
  as model prior for deep learning,'' in {\em International Conference on
  Learning Representations}, 2019.

\bibitem{cranmer2020lagrangian}
M.~Cranmer, S.~Greydanus, S.~Hoyer, P.~Battaglia, D.~Spergel, and S.~Ho,
  ``Lagrangian neural networks,'' in {\em ICLR 2020 Workshop on Integration of
  Deep Neural Models and Differential Equations}, 2020.

\bibitem{raissi2019physics}
M.~Raissi, P.~Perdikaris, and G.~E. Karniadakis, ``Physics-informed neural
  networks: A deep learning framework for solving forward and inverse problems
  involving nonlinear partial differential equations,'' {\em Journal of
  Computational Physics}, vol.~378, pp.~686--707, 2019.

\bibitem{rackauckas2020universal}
C.~Rackauckas, Y.~Ma, J.~Martensen, C.~Warner, K.~Zubov, R.~Supekar,
  D.~Skinner, A.~Ramadhan, and A.~Edelman, ``Universal differential equations
  for scientific machine learning,'' {\em arXiv preprint arXiv:2001.04385},
  2020.

\bibitem{zhan2006neural}
R.~Zhan and J.~Wan, ``Neural network-aided adaptive unscented kalman filter for
  nonlinear state estimation,'' {\em IEEE Signal Processing Letters}, vol.~13,
  no.~7, pp.~445--448, 2006.

\bibitem{tagliabue2020touch}
A.~Tagliabue, A.~Paris, S.~Kim, R.~Kubicek, S.~Bergbreiter, and J.~P. How,
  ``Touch the wind: Simultaneous airflow, drag and interaction sensing on a
  multirotor,'' in {\em 2020 IEEE/RSJ International Conference on Intelligent
  Robots and Systems (IROS)}, pp.~1645--1652, IEEE, 2020.

\bibitem{noroozi2008generation}
N.~Noroozi, P.~Karimaghaee, F.~Safaei, and H.~Javadi, ``Generation of lyapunov
  functions by neural networks,'' in {\em Proceedings of the World Congress on
  Engineering}, vol.~2008, 2008.

\bibitem{yeung2019learning}
E.~Yeung, S.~Kundu, and N.~Hodas, ``Learning deep neural network
  representations for koopman operators of nonlinear dynamical systems,'' in
  {\em 2019 American Control Conference (ACC)}, pp.~4832--4839, IEEE, 2019.

\bibitem{bansal2020deepreach}
S.~Bansal and C.~Tomlin, ``{DeepReach}: A deep learning approach to
  high-dimensional reachability,'' in {\em IEEE International Conference on
  Robotics and Automation (ICRA)}, 2021.

\bibitem{chen2021learning}
S.~Chen, M.~Fazlyab, M.~Morari, G.~J. Pappas, and V.~M. Preciado, ``Learning
  lyapunov functions for hybrid systems,'' in {\em Proceedings of the 24th
  International Conference on Hybrid Systems: Computation and Control},
  pp.~1--11, 2021.

\bibitem{luckcuck2019formal}
M.~Luckcuck, M.~Farrell, L.~A. Dennis, C.~Dixon, and M.~Fisher, ``Formal
  specification and verification of autonomous robotic systems: A survey,''
  {\em ACM Computing Surveys (CSUR)}, vol.~52, no.~5, pp.~1--41, 2019.

\bibitem{garoche2019formal}
P.-L. Garoche, {\em Formal verification of control system software}, vol.~67.
\newblock Princeton University Press, 2019.

\bibitem{tabuada2009verification}
P.~Tabuada, {\em Verification and control of hybrid systems: a symbolic
  approach}.
\newblock Springer Science \& Business Media, 2009.

\bibitem{zheng2015perceptions}
X.~Zheng, C.~Julien, M.~Kim, and S.~Khurshid, ``Perceptions on the state of the
  art in verification and validation in cyber-physical systems,'' {\em IEEE
  Systems Journal}, vol.~11, no.~4, pp.~2614--2627, 2015.

\bibitem{tomlin2000game}
C.~J. Tomlin, J.~Lygeros, and S.~S. Sastry, ``A game theoretic approach to
  controller design for hybrid systems,'' {\em Proceedings of the IEEE},
  vol.~88, no.~7, pp.~949--970, 2000.

\bibitem{frehse2011spaceex}
G.~Frehse, C.~Le~Guernic, A.~Donz{\'e}, S.~Cotton, R.~Ray, O.~Lebeltel,
  R.~Ripado, A.~Girard, T.~Dang, and O.~Maler, ``Spaceex: Scalable verification
  of hybrid systems,'' in {\em International Conference on Computer Aided
  Verification}, pp.~379--395, Springer, 2011.

\bibitem{chen2013flow}
X.~Chen, E.~{\'A}brah{\'a}m, and S.~Sankaranarayanan, ``Flow*: An analyzer for
  non-linear hybrid systems,'' in {\em International Conference on Computer
  Aided Verification}, pp.~258--263, Springer, 2013.

\bibitem{althoff2015introduction}
M.~Althoff, ``An introduction to cora 2015,'' in {\em Proc. of the Workshop on
  Applied Verification for Continuous and Hybrid Systems}, 2015.

\bibitem{duggirala2015c2e2}
P.~S. Duggirala, S.~Mitra, M.~Viswanathan, and M.~Potok, ``C2e2: A verification
  tool for stateflow models,'' in {\em International Conference on Tools and
  Algorithms for the Construction and Analysis of Systems}, pp.~68--82,
  Springer, 2015.

\bibitem{fan2016automatic}
C.~Fan, B.~Qi, S.~Mitra, M.~Viswanathan, and P.~S. Duggirala, ``Automatic
  reachability analysis for nonlinear hybrid models with c2e2,'' in {\em
  International Conference on Computer Aided Verification}, pp.~531--538,
  Springer, 2016.

\bibitem{papachristodoulou2002construction}
A.~Papachristodoulou and S.~Prajna, ``On the construction of lyapunov functions
  using the sum of squares decomposition,'' in {\em Proceedings of the 41st
  IEEE Conference on Decision and Control, 2002.}, vol.~3, pp.~3482--3487,
  IEEE, 2002.

\bibitem{ames2016control}
A.~D. Ames, X.~Xu, J.~W. Grizzle, and P.~Tabuada, ``Control barrier function
  based quadratic programs for safety critical systems,'' {\em IEEE
  Transactions on Automatic Control}, vol.~62, no.~8, pp.~3861--3876, 2016.

\bibitem{Ehlers_2017}
R.~Ehlers, ``Formal verification of piece-wise linear feed-forward neural
  networks,'' in {\em ATVA}, 2017.

\bibitem{Katz_2017}
G.~Katz, C.~W. Barrett, D.~L. Dill, K.~Julian, and M.~J. Kochenderfer,
  ``Reluplex: An efficient {SMT} solver for verifying deep neural networks,''
  in {\em Computer Aided Verification - 29th International Conference, {CAV}
  2017, Heidelberg, Germany, July 24-28, 2017, Proceedings, Part {I}},
  pp.~97--117, 2017.

\bibitem{Huang_2017b}
X.~Huang, M.~Kwiatkowska, S.~Wang, and M.~Wu, ``Safety verification of deep
  neural networks,'' in {\em Computer Aided Verification} (R.~Majumdar and
  V.~Kun{\v{c}}ak, eds.), (Cham), pp.~3--29, Springer International Publishing,
  2017.

\bibitem{Lomuscio_2017}
A.~Lomuscio and L.~Maganti, ``An approach to reachability analysis for
  feed-forward relu neural networks,'' {\em CoRR}, vol.~abs/1706.07351, 2017.

\bibitem{Tjeng_2019}
V.~Tjeng, K.~Y. Xiao, and R.~Tedrake, ``Evaluating robustness of neural
  networks with mixed integer programming,'' in {\em International Conference
  on Learning Representations (ICLR)}, 2019.

\bibitem{Gehr_2018}
T.~{Gehr}, M.~{Mirman}, D.~{Drachsler-Cohen}, P.~{Tsankov}, S.~{Chaudhuri}, and
  M.~{Vechev}, ``Ai2: Safety and robustness certification of neural networks
  with abstract interpretation,'' in {\em 2018 IEEE Symposium on Security and
  Privacy (SP)}, pp.~3--18, May 2018.

\bibitem{gowal2018effectiveness}
S.~Gowal, K.~Dvijotham, R.~Stanforth, R.~Bunel, C.~Qin, J.~Uesato,
  R.~Arandjelovic, T.~Mann, and P.~Kohli, ``On the effectiveness of interval
  bound propagation for training verifiably robust models,'' {\em arXiv
  preprint arXiv:1810.12715}, 2018.

\bibitem{Weng_2018}
T.~Weng, H.~Zhang, H.~Chen, Z.~Song, C.~Hsieh, L.~Daniel, D.~Boning, and
  I.~Dhillon, ``Towards fast computation of certified robustness for relu
  networks,'' in {\em International Conference on Machine Learning (ICML)},
  2018.

\bibitem{singh2018fast}
G.~Singh, T.~Gehr, M.~Mirman, M.~P{\"u}schel, and M.~Vechev, ``Fast and
  effective robustness certification,'' in {\em Advances in Neural Information
  Processing Systems}, pp.~10802--10813, 2018.

\bibitem{zhang2018efficient}
H.~Zhang, T.-W. Weng, P.-Y. Chen, C.-J. Hsieh, and L.~Daniel, ``Efficient
  neural network robustness certification with general activation functions,''
  in {\em Advances in neural information processing systems}, pp.~4939--4948,
  2018.

\bibitem{Wong_2018}
E.~Wong and J.~Z. Kolter, ``Provable defenses against adversarial examples via
  the convex outer adversarial polytope,'' in {\em {ICML}}, vol.~80 of {\em
  Proceedings of Machine Learning Research}, pp.~5283--5292, 2018.

\bibitem{raghunathan2018certified}
A.~Raghunathan, J.~Steinhardt, and P.~Liang, ``Certified defenses against
  adversarial examples,'' in {\em International Conference on Learning
  Representations}, 2018.

\bibitem{fazlyab2020safety}
M.~Fazlyab, M.~Morari, and G.~J. Pappas, ``Safety verification and robustness
  analysis of neural networks via quadratic constraints and semidefinite
  programming,'' {\em IEEE Transactions on Automatic Control}, 2020.

\bibitem{xu2020automatic}
K.~Xu, Z.~Shi, H.~Zhang, Y.~Wang, K.-W. Chang, M.~Huang, B.~Kailkhura, X.~Lin,
  and C.-J. Hsieh, ``Automatic perturbation analysis for scalable certified
  robustness and beyond,'' {\em Advances in Neural Information Processing
  Systems}, vol.~33, 2020.

\bibitem{tran2020verification}
H.-D. Tran, W.~Xiang, and T.~T. Johnson, ``Verification approaches for
  learning-enabled autonomous cyber-physical systems,'' {\em IEEE Design \&
  Test}, 2020.

\bibitem{xiang2018verification}
W.~Xiang, P.~Musau, A.~A. Wild, D.~M. Lopez, N.~Hamilton, X.~Yang,
  J.~Rosenfeld, and T.~T. Johnson, ``Verification for machine learning,
  autonomy, and neural networks survey,'' {\em arXiv preprint
  arXiv:1810.01989}, 2018.

\bibitem{tanaka1996approach}
K.~Tanaka, ``An approach to stability criteria of neural-network control
  systems,'' {\em IEEE Transactions on Neural Networks}, vol.~7, no.~3,
  pp.~629--642, 1996.

\bibitem{yin2021stability}
H.~Yin, P.~Seiler, and M.~Arcak, ``Stability analysis using quadratic
  constraints for systems with neural network controllers,'' {\em IEEE
  Transactions on Automatic Control}, 2021.

\bibitem{dutta2019reachability}
S.~Dutta, X.~Chen, and S.~Sankaranarayanan, ``Reachability analysis for neural
  feedback systems using regressive polynomial rule inference,'' in {\em
  Proceedings of the 22nd ACM International Conference on Hybrid Systems:
  Computation and Control}, pp.~157--168, 2019.

\bibitem{huang2019reachnn}
C.~Huang, J.~Fan, W.~Li, X.~Chen, and Q.~Zhu, ``Reachnn: Reachability analysis
  of neural-network controlled systems,'' {\em ACM Transactions on Embedded
  Computing Systems (TECS)}, vol.~18, no.~5s, pp.~1--22, 2019.

\bibitem{fan2020reachnn}
J.~Fan, C.~Huang, X.~Chen, W.~Li, and Q.~Zhu, ``Reachnn*: A tool for
  reachability analysis of neural-network controlled systems,'' in {\em
  International Symposium on Automated Technology for Verification and
  Analysis}, pp.~537--542, Springer, 2020.

\bibitem{ivanov2019verisig}
R.~Ivanov, J.~Weimer, R.~Alur, G.~J. Pappas, and I.~Lee, ``Verisig: verifying
  safety properties of hybrid systems with neural network controllers,'' in
  {\em Proceedings of the 22nd ACM International Conference on Hybrid Systems:
  Computation and Control}, pp.~169--178, 2019.

\bibitem{xiang2020reachable}
W.~Xiang, H.-D. Tran, X.~Yang, and T.~T. Johnson, ``Reachable set estimation
  for neural network control systems: A simulation-guided approach,'' {\em IEEE
  Transactions on Neural Networks and Learning Systems}, 2020.

\bibitem{hu2020reach}
H.~Hu, M.~Fazlyab, M.~Morari, and G.~J. Pappas, ``Reach-sdp: Reachability
  analysis of closed-loop systems with neural network controllers via
  semidefinite programming,'' in {\em 59th IEEE Conference on Decision and
  Control}, 2020.

\bibitem{yang2019efficient}
G.~Yang, G.~Qian, P.~Lv, and H.~Li, ``Efficient verification of control systems
  with neural network controllers,'' in {\em Proceedings of the 3rd
  International Conference on Vision, Image and Signal Processing}, pp.~1--7,
  2019.

\bibitem{vincent2021reachable}
J.~A. Vincent and M.~Schwager, ``Reachable polyhedral marching (rpm): A safety
  verification algorithm for robotic systems with deep neural network
  components,'' 2021.

\bibitem{gpu_implementation_crown}
H.~Zhang, H.~Chen, C.~Xiao, S.~Gowal, R.~Stanforth, B.~Li, D.~Boning, and C.-J.
  Hsieh, ``Crown-ibp: Towards stable and efficient training of verifiably
  robust neural networks.'' \url{https://github.com/huanzhang12/CROWN-IBP},
  2020.

\bibitem{salman2019convex}
H.~Salman, G.~Yang, H.~Zhang, C.-J. Hsieh, and P.~Zhang, ``A convex relaxation
  barrier to tight robustness verification of neural networks,'' in {\em
  Advances in Neural Information Processing Systems}, pp.~9835--9846, 2019.

\bibitem{katz2017reluplex}
G.~Katz, C.~Barrett, D.~L. Dill, K.~Julian, and M.~J. Kochenderfer, ``Reluplex:
  An efficient smt solver for verifying deep neural networks,'' in {\em
  International Conference on Computer Aided Verification}, pp.~97--117,
  Springer, 2017.

\bibitem{weng2018towards}
L.~Weng, H.~Zhang, H.~Chen, Z.~Song, C.-J. Hsieh, L.~Daniel, D.~Boning, and
  I.~Dhillon, ``Towards fast computation of certified robustness for relu
  networks,'' in {\em International Conference on Machine Learning},
  pp.~5276--5285, PMLR, 2018.

\bibitem{dathathri2020enabling}
S.~Dathathri, K.~Dvijotham, A.~Kurakin, A.~Raghunathan, J.~Uesato, R.~Bunel,
  S.~Shankar, J.~Steinhardt, I.~Goodfellow, P.~Liang, {\em et~al.}, ``Enabling
  certification of verification-agnostic networks via memory-efficient
  semidefinite programming,'' in {\em Advances in Neural Information Processing
  Systems}, vol.~33, pp.~5318--5331, 2020.

\bibitem{singh2019beyond}
G.~Singh, R.~Ganvir, M.~P{\"u}schel, and M.~Vechev, ``Beyond the single neuron
  convex barrier for neural network certification,'' in {\em Advances in Neural
  Information Processing Systems}, pp.~15098--15109, 2019.

\bibitem{tjandraatmadja2020convex}
C.~Tjandraatmadja, R.~Anderson, J.~Huchette, W.~Ma, K.~K. PATEL, and J.~P.
  Vielma, ``The convex relaxation barrier, revisited: Tightened single-neuron
  relaxations for neural network verification,'' {\em Advances in Neural
  Information Processing Systems}, vol.~33, 2020.

\bibitem{muller2021prima}
M.~N. M{\"u}ller, G.~Makarchuk, G.~Singh, M.~P{\"u}schel, and M.~Vechev,
  ``Prima: Precise and general neural network certification via multi-neuron
  convex relaxations,'' {\em arXiv preprint arXiv:2103.03638}, 2021.

\bibitem{anderson2020tightened}
B.~G. Anderson, Z.~Ma, J.~Li, and S.~Sojoudi, ``Tightened convex relaxations
  for neural network robustness certification,'' in {\em 2020 59th IEEE
  Conference on Decision and Control (CDC)}, pp.~2190--2197, IEEE, 2020.

\bibitem{xiang2018output}
W.~Xiang, H.-D. Tran, and T.~T. Johnson, ``Output reachable set estimation and
  verification for multilayer neural networks,'' {\em IEEE transactions on
  neural networks and learning systems}, vol.~29, no.~11, pp.~5777--5783, 2018.

\bibitem{wang2018formal}
S.~Wang, K.~Pei, J.~Whitehouse, J.~Yang, and S.~Jana, ``Formal security
  analysis of neural networks using symbolic intervals,'' in {\em 27th
  $\{$USENIX$\}$ Security Symposium ($\{$USENIX$\}$ Security 18)},
  pp.~1599--1614, 2018.

\bibitem{rubies2019fast}
V.~Rubies-Royo, R.~Calandra, D.~M. Stipanovic, and C.~Tomlin, ``Fast neural
  network verification via shadow prices,'' {\em arXiv preprint
  arXiv:1902.07247}, 2019.

\bibitem{fazlyab2019probabilistic}
M.~Fazlyab, M.~Morari, and G.~J. Pappas, ``Probabilistic verification and
  reachability analysis of neural networks via semidefinite programming,'' in
  {\em 2019 IEEE 58th Conference on Decision and Control (CDC)},
  pp.~2726--2731, IEEE, 2019.

\bibitem{diamond2016cvxpy}
S.~Diamond and S.~Boyd, ``Cvxpy: A python-embedded modeling language for convex
  optimization,'' {\em The Journal of Machine Learning Research}, vol.~17,
  no.~1, pp.~2909--2913, 2016.

\bibitem{Mnih_2015}
V.~Mnih, K.~Kavukcuoglu, D.~Silver, A.~A. Rusu, J.~Veness, M.~G. Bellemare,
  A.~Graves, M.~Riedmiller, A.~K. Fidjeland, G.~Ostrovski, S.~Petersen,
  C.~Beattie, A.~Sadik, I.~Antonoglou, H.~King, D.~Kumaran, D.~Wierstra,
  S.~Legg, and D.~Hassabis, ``Human-level control through deep reinforcement
  learning,'' in {\em Nature}, vol.~518, Nature Publishing Group, a division of
  Macmillan Publishers Limited., 2015.

\bibitem{ilahi2021challenges}
I.~Ilahi, M.~Usama, J.~Qadir, M.~U. Janjua, A.~Al-Fuqaha, D.~T. Huang, and
  D.~Niyato, ``Challenges and countermeasures for adversarial attacks on deep
  reinforcement learning,'' {\em IEEE Transactions on Artificial Intelligence},
  2021.

\bibitem{haarnoja2018soft}
T.~Haarnoja, A.~Zhou, P.~Abbeel, and S.~Levine, ``Soft actor-critic: Off-policy
  maximum entropy deep reinforcement learning with a stochastic actor,'' in
  {\em International conference on machine learning}, pp.~1861--1870, PMLR,
  2018.

\bibitem{christodoulou2019soft}
P.~Christodoulou, ``Soft actor-critic for discrete action settings,'' {\em
  arXiv preprint arXiv:1910.07207}, 2019.

\end{thebibliography}
